\documentclass{article}
\usepackage{iclr2027_conference,times}

\usepackage{placeins}
\usepackage[utf8]{inputenc}
\usepackage[T1]{fontenc}
\usepackage{hyperref}
\hypersetup{hidelinks,pdftitle={Language Model Augmented Decision Transformer for Multi-Task Offline Reinforcement Learning},pdfauthor={Xinbo Zhao, Yuhang Liu, Xin Zhang, Lei Le, Zhaohan Xi, Yu Yang, Yingxue Zhang}}
\usepackage{url}
\usepackage{booktabs}
\usepackage{amsfonts}
\usepackage{nicefrac}
\usepackage{microtype}
\usepackage{xcolor}
\usepackage{enumitem}

\usepackage{graphicx}
\usepackage{subcaption}
\usepackage{multirow}
\usepackage{bm}
\usepackage{amsmath}
\usepackage{amssymb}
\usepackage{mathtools}
\usepackage{amsthm}
\usepackage{xspace}
\usepackage{algorithm}
\usepackage{algorithmic}
\usepackage[capitalize,noabbrev]{cleveref}

\theoremstyle{plain}

\theoremstyle{definition}

\theoremstyle{remark}

\makeatletter
\DeclareRobustCommand\onedot{\futurelet\@let@token\@onedot}
\def\@onedot{\ifx\@let@token.\else.\null\fi\xspace}

\makeatother

\title{Language Model Augmented Decision\\
Transformer for Multi-Task Offline\\
Reinforcement Learning}

\author{%
\begin{minipage}[t]{\dimexpr\textwidth-2\tabcolsep\relax}
\centering
\normalfont\normalsize
\textbf{Xinbo Zhao}$^{1}$, \textbf{Yuhang Liu}$^{1}$, \textbf{Xin Zhang}$^{2}$, \textbf{Lei Le}$^{3}$,\\
\textbf{Zhaohan Xi}$^{1}$, \textbf{Yu Yang}$^{4}$, \textbf{Yingxue Zhang}$^{1}$\\[0.4em]
\small
$^{1}$State University of New York at Binghamton\\
$^{2}$San Diego State University\quad $^{3}$LinkedIn\quad $^{4}$Lehigh University\\[0.35em]
\footnotesize
\texttt{\{xzhao9,yliu41,zxi1,yzhang42\}@binghamton.edu}\\
\texttt{xzhang19@sdsu.edu}\quad\texttt{some.other.lei@gmail.com}\quad\texttt{yuyang@lehigh.edu}
\end{minipage}}

\iclrfinalcopy

\begin{document}

\maketitle
\lhead{Preprint}
\raggedbottom

\begin{abstract}
Offline multi-task reinforcement learning (MTRL) aims to learn a unified policy from heterogeneous offline datasets that generalizes across tasks. However, existing offline MTRL methods often struggle with two fundamental issues: severe per-task data scarcity and limited cross-task generalization.
In this paper, we propose \underline{La}nguage-\underline{M}odel \underline{A}ugmented \underline{D}ecision \underline{T}ransformer (LaMA-DT), a unified sequence-modeling framework that leverages a single large language model (LLM) for both trajectory augmentation and multi-task policy learning. LaMA-DT uses natural-language task descriptions to guide task-aligned trajectory augmentation through a set of supervised generation and alignment objectives, and incorporates a scoring-and-filtering mechanism to retain high-quality augmented trajectories. The same LLM is then repurposed as a language-conditioned Decision Transformer and fine-tuned on a mixture of real and augmented data. %
Experiments on Meta-World and MuJoCo multi-task benchmarks demonstrate that LaMA-DT consistently outperforms competitive baselines, particularly in low-data regimes, showing that language-guided data augmentation substantially improves cross-task generalization.
\end{abstract}

\section{Introduction}

Offline reinforcement learning (RL)~\citep{levine2020offline} enables agents to learn high-quality decision policies directly from static datasets. To extend offline RL to more realistic applications, it is important to handle multi-task settings, where an agent must solve multiple related tasks that share state and action spaces but differ in goals, rewards, or context. This motivates offline multi-task RL (MTRL)~\citep{yu2020meta}, which learns from heterogeneous offline datasets and aims to produce a single policy that can perform well across tasks.
In practice, offline MTRL is limited by two tightly coupled challenges. First, \textit{per-task data is often scarce:} each task may have only a small and fragmented dataset, since collecting trajectories for many tasks is expensive and often infeasible. Second, \textit{cross-task generalization is difficult:} tasks vary in objectives, dynamics, and observation distributions, leading to substantial distribution shifts that hinder learning a single policy across tasks, especially when limited state–action coverage further amplifies these shifts.

Existing offline MTRL methods offer partial solutions, but typically target only one challenge or rely on strong data assumptions. For improving data use across tasks,
CDS~\citep{yu2021conservative} transfers transitions from auxiliary tasks to a target task based on estimated usefulness. While effective when abundant and closely related auxiliary data are available, CDS struggles with cross-task generalization, as it requires training a separate model for each target task and degrades when auxiliary data are scarce or poorly matched. SRTD~\citep{yoo2022skills} learns a shared latent skill space and performs data augmentation via imagined rollouts; however, learning a stable and transferable latent representation generally requires sufficient task-specific data. When per-task datasets are limited, the learned latent space becomes unreliable, reducing the effectiveness of the generated rollouts.
A separate line of work builds on Decision Transformers (DTs)~\citep{chen2021decision} and adapts them to multi-task learning to improve generalization through shared sequence modeling. For example, HarmoDT~\citep{hu2024harmodt} improves parameter sharing by reducing gradient conflicts across tasks, and SoCo-DT~\citep{wang2025socodt} refines sharing via Fisher-information--guided soft masks. However, these DT-based MTRL methods still typically rely on having enough trajectories per task to learn transferable sequence representations; when each task has limited data, performance can drop sharply due to insufficient coverage and overfitting to narrow task-specific patterns.
These gaps naturally raise a question: \emph{In offline MTRL under \textbf{low-data regimes} (i.e., with scarce task-specific data), how can we expand effective data coverage while achieving robust cross-task generalization?}

A key observation is that, although per-task trajectories are limited, high-level task descriptions are almost always available. Such descriptions convey information such as the task name, goal, object identity, or expertise level, which can be succinctly expressed in natural language. This motivates our approach: use natural language as a task-specification medium to guide trajectory augmentation via large language models (LLMs), thereby improving multi-task policy learning in data-scarce scenarios.
However, since LLMs are not pretrained on trajectory data, realizing this idea requires ensuring that augmented trajectories both remain faithful to limited real data while expanding state–action coverage, and stay aligned with task descriptions to prevent cross-task drift.

In this paper, we propose \underline{La}nguage-\underline{M}odel \underline{A}ugmented \underline{D}ecision \underline{T}ransformer (LaMA-DT) for offline MTRL, which employs a single LLM for both trajectory augmentation and multi-task policy learning.
To enable high-quality, task-aligned trajectory generation, the LLM is first fine-tuned across tasks using four carefully designed supervised objectives that jointly capture environmental dynamics and temporal dependencies in sequential decision-making.
To further ensure semantic alignment between natural-language task descriptions and generated trajectories, LaMA-DT incorporates two additional task–trajectory alignment objectives.
All six objectives are jointly optimized under a dedicated multi-objective fine-tuning strategy tailored for LLM-based trajectory learning.
To ensure that only high-quality augmented trajectories are used for policy learning, we introduce a likelihood-based scoring-and-filtering mechanism that evaluates generated trajectories across all objectives and discards low-quality or task-misaligned samples.
Finally, the same LLM is repurposed as a DT—a natural choice for sequence modeling—conditioned on natural-language task descriptions and fine-tuned on a mixture of real and high-quality augmented trajectories, resulting in robust cross-task generalization.
Key contributions are summarized as follows:
\begin{itemize}[nosep, leftmargin=*]
    \item We propose LaMA-DT, an LLM-based offline MTRL framework that unifies trajectory augmentation and multi-task policy learning. It addresses data scarcity by leveraging natural-language task descriptions to guide task-aligned trajectory augmentation via a set of objectives that capture task-specific data distributions and enforce semantic alignment.%
    
    \item 
    To enable cross-task generalization in offline MTRL, LaMA-DT introduces an LLM-based policy learning mechanism that combines scored and filtered high-quality augmented trajectories with original data, conditioned on task descriptions, to fine-tune a DT initialized from the same LLM, yielding improved cross-task generalization.
    
    \item LaMA-DT demonstrates consistent improvements over competitive baselines on Meta-World and MuJoCo multi-task benchmarks, with particularly strong gains in \textbf{low-data regimes}, highlighting that effective data augmentation substantially enhances multi-task learning performance\footnote{Code available at \url{https://anonymous.4open.science/r/o33zlamadt-6587}.}.
\end{itemize}

\section{Preliminaries}

\paragraph{Offline Multi-Task Reinforcement Learning (MTRL).}
An RL problem is typically modeled as a Markov decision process (MDP), defined by the tuple 
$(\mathcal{S}, \mathcal{A}, P, r, \gamma)$, 
where $\mathcal{S}$ is the state space, $\mathcal{A}$ is the action space, 
$P(s' \mid s,a)$ is the transition distribution, 
$r(s,a)$ is the reward function, 
and $\gamma \in [0,1)$ is the discount factor. 
The goal is to learn a policy $\pi(a \mid s)$ that maximizes the expected discounted return: 
$J(\pi) = \mathbb{E}_{\pi} \left[ \sum_{t=0}^{\infty} \gamma^t r(s_t, a_t) \right].$ 
In offline RL, the agent must learn solely from a fixed dataset of trajectories 
$\mathcal{D} = \{\tau_j\}_{j=1}^{N}$ 
collected by an unknown behavior policy, where each trajectory 
$\tau_j = (s_0, a_0, r_0, \dots, s_T, a_T, r_T)$.
Offline MTRL extends the offline RL setting to a collection of tasks 
$\mathcal{T} = \{ \mathcal{T}_i \}_{i=1}^{M}$,
where each task $\mathcal{T}_i$ corresponds to its own MDP 
$(\mathcal{S}, \mathcal{A}, P^{\mathcal{T}_i}, r^{\mathcal{T}_i}, \gamma)$
and is associated with an offline dataset 
$\mathcal{D}_i = \{ \tau_j^{\,i} \}$.
Tasks may differ in their reward functions and transition dynamics, 
but share a common state and action space.
The objective is to learn a single policy that 
performs well across this fixed set of tasks. Formally, the optimal policy 
is defined as: 
$\pi^{*}
= \arg\max_{\pi} \;
\frac{1}{M} \sum_{i=1}^{M} 
\mathbb{E}_{\pi} 
\left[ \sum_{t=0}^{\infty} \gamma^{t} r^{\mathcal{T}_i}_{t} \right].$

\noindent\textbf{Decision Transformer (DT).}
DT~\citep{chen2021decision} formulates offline RL as a return-conditioned sequence modeling problem using a transformer architecture, learning policies directly from offline trajectories by autoregressively predicting actions.
In DT, each trajectory is represented as a sequence of tokens 
$(\hat{R_t}, s_t, a_t)$, 
where 
$\hat{R_t} = \sum_{t'=t}^{T} r_{t'}$ 
is the return-to-go (RTG) from timestep $t$. The RTG serves as a conditioning signal, encouraging the model to imitate high-return behaviors present in the dataset.
The model is trained to predict the next action $a'_t$ given the history and current state, with the ground-truth action $a_t$ as supervision.

\section{Methodology}
\begin{figure}[t]
\centering
\includegraphics[width=1\textwidth]{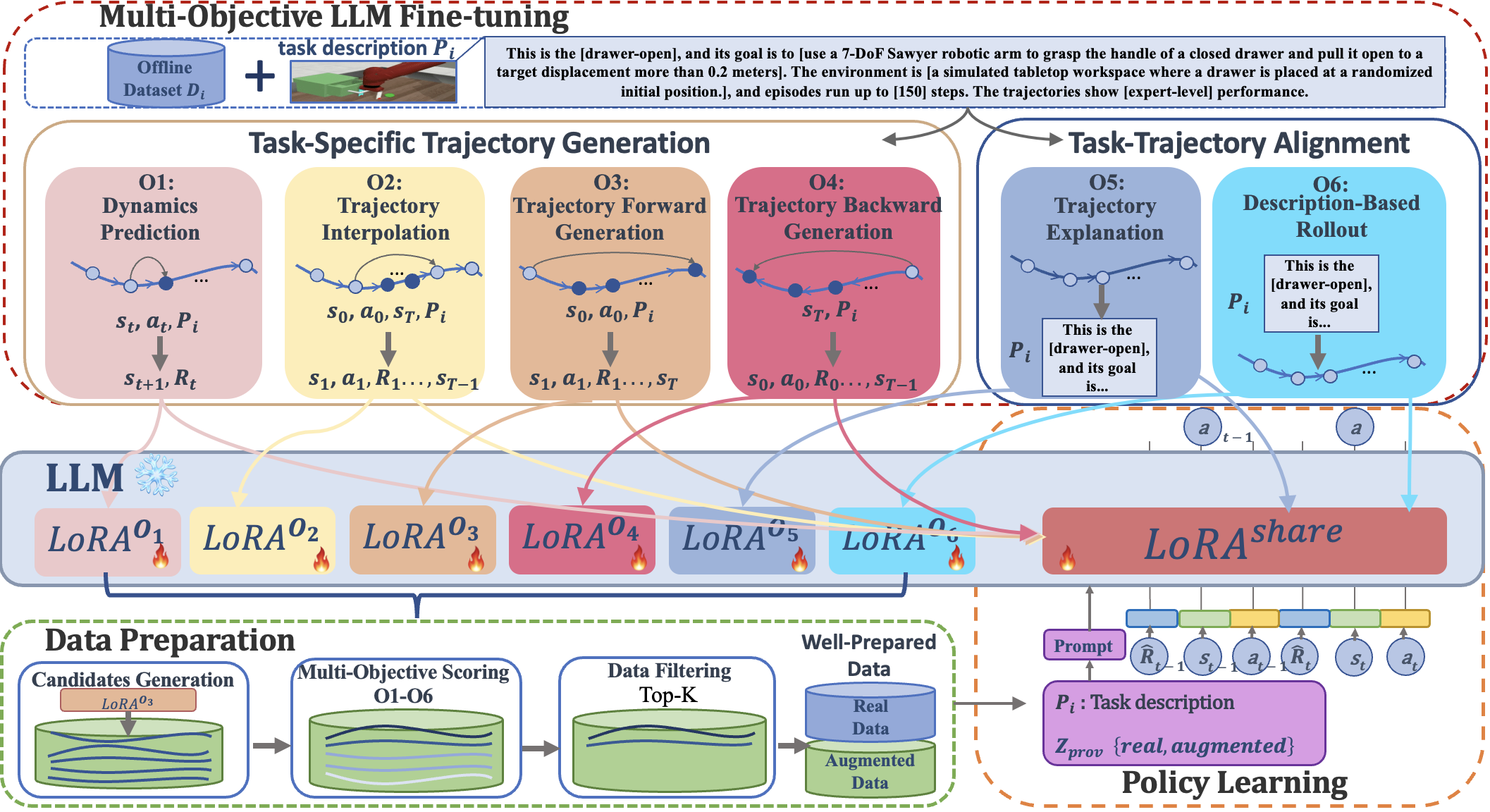}
\caption{
LaMA-DT consists of three components:
(i) Multi-Objective Fine-Tuning. We adapt the LLM to generate meaningful, task-specific trajectories from limited offline data and natural-language task descriptions by training objective-specific LoRA adapters together with a shared LoRA adapter, enabling both trajectory generation and task--trajectory alignment.
(ii) Data Preparation. Candidate trajectories are generated per task using $\text{O}_3$, scored using aggregated multi-objective likelihoods, and the top-$K$ trajectories are retained and merged with the original offline data.
(iii) Multi-Task Policy Learning. We repurpose the LLM and shared LoRA adapter as a prompt-conditioned DT, trained on the combined real and augmented dataset, with task descriptions $\mathbf{p}_i$ and a provenance token $z_{\text{prov}}$.
}
\label{fig:framework}
\end{figure}

In this section, we present Language-Model Augmented Decision Transformer (LaMA-DT) for offline MTRL, which employs a single LLM for both trajectory augmentation and multi-task policy learning. Fig.~\ref{fig:framework} illustrates the framework overview.

\subsection{Task Descriptions in Offline MTRL}
A central difficulty in offline MTRL is data scarcity: per-task datasets are often small, making it difficult to learn reliable policies that generalize across tasks.
Although trajectory data are scarce, rich task-level metadata are almost always available. Such metadata—including task names, goals, relevant objects, environmental conditions, and typical performance—can be naturally expressed in natural language.
This observation motivates the use of natural-language task descriptions to guide an LLM in generating additional, task-aligned trajectories, leveraging its strong sequence-modeling and reasoning capabilities to expand effective data coverage and thus mitigate data scarcity.
To operationalize this idea, we construct a standardized natural-language description for each task $\mathcal{T}_i \in \mathcal{T}$ using a simple template:
\texttt{``This is the [task\_name], and its goal is to [task\_goal].
The environment is [environment\_description].
and episodes run up to [max\_episode\_length] steps. The trajectories show [expertise\_level] performance.''}

Each placeholder is deterministically filled by metadata extracted from task $\mathcal{T}_i$. %
Formally, our training data consist of pairs of task descriptions and their associated trajectories,
$\bigl\{\,(\mathbf{p}_i,\; \mathcal{D}_i = \{\tau_j^{\,i}\})\,\bigr\}_{i=1}^{M},$
where $\mathbf{p}_i$ denotes the natural-language description of task $\mathcal{T}_i$.

\subsection{Task-Aligned Trajectory Augmentation with LLMs}
\label{sec:augmentation}

\subsubsection{Task-Specific Trajectory Generation}
\label{sec:Trajectory Generation}
We design four trajectory-generation objectives ($\text{O}_1$–$\text{O}_4$) to fine-tune the LLM, enabling it to capture environmental dynamics, temporal dependencies in sequential decision-making, and task-specific data distributions, thereby generating plausible trajectories that resemble the available data while effectively expanding state–action coverage.

\textbf{Dynamics Learning Objective.}
We first equip the LLM with an explicit understanding of environment dynamics to ensure physically grounded trajectory generation via $\text{O}_1$:

\textit{Objective 1: Dynamics Prediction ($\text{O}_1$).}
Accurate modeling of dynamics is essential to prevent physically implausible trajectories (\emph{e.g.}, a robot arm violating joint limits or a locomotion agent defying gravity).
$\text{O}_1$ fine-tunes the LLM to predict the next state $s'$ given the current state $s$ and action $a$ in a supervised manner, enabling it to capture the transition dynamics.
For each task $\mathcal{T}_i$, the LLM receives its task description $\mathbf{p}_i$ along with the prompt:
\texttt{``Predict the next state given the current state and action $(s, a)$.''}

\textbf{Sequential Decision Learning Objectives.}
Beyond modeling environmental dynamics, we design $\text{O}_2$--$\text{O}_4$ to capture temporal dependencies in sequential decision-making across different tasks from complementary perspectives, while also reinforcing the dynamics learned in $\text{O}_1$:

\textit{Objective 2: Trajectory Interpolation ($\text{O}_2$).}
$\text{O}_2$ is designed to generate plausible intermediate states, actions, and rewards between two observed points along a trajectory. This objective encourages the LLM to preserve local temporal consistency, ensuring that short-horizon transitions remain coherent within the overall trajectory and thereby reducing error accumulation in long-horizon rollouts.
In practice, we randomly select two states from a real trajectory as the start and end states, and use the ground-truth subtrajectory between them as supervision.
For $\text{O}_2$, the LLM is provided with the task description $\mathbf{p}_i$ for task $\mathcal{T}_i$ along with the prompt:
\texttt{``Fill in the missing part of the trajectory given the start state and action $(s_0, a_0)$ and the final state $s_T$.''}

\textit{Objective 3: Trajectory Forward Generation ($\text{O}_3$).}
Forward generation is a core capability for trajectory modeling: given an initial state–action pair, the model should be able to generate a plausible rollout over time.
$\text{O}_3$ aims to capture long-horizon temporal dependencies by encouraging the LLM to maintain consistency throughout the trajectory in the forward direction.
Specifically, $\text{O}_3$ fine-tunes the LLM in a supervised manner to generate complete trajectories starting from $(s_0, a_0)$, using trajectories from $\mathcal{D}_i$ as ground-truth supervision.
For $\text{O}_3$, the LLM is provided with the task description $\mathbf{p}_i$ for task $\mathcal{T}_i$ along with the prompt:
\texttt{``Generate the whole trajectory until termination given the start state and action $(s_0, a_0)$.''}

\textit{Objective 4: Trajectory Backward Generation ($\text{O}_4$).}
Human learning often relies on bidirectional reasoning: we not only predict future consequences from a current state but also reason backward from a desired outcome.
A similar capability is valuable for trajectory generation.
Relying solely on forward prediction may lead to error accumulation or trajectories that diverge from the intended task goal—especially in multi-task settings where tasks may share similar local dynamics but differ substantially in their global objectives.
$\text{O}_4$ is designed to address this issue by fine-tuning the LLM to reconstruct trajectories backward from a given final state with supervision.
For $\text{O}_4$, the LLM is provided with the task description $\mathbf{p}_i$ for task $\mathcal{T}_i$ along with the prompt:
\texttt{``Reconstruct the trajectory backwards from the final state $s_T$.''}

\subsubsection{Task–Trajectory Alignment}
\label{sec:alignment}

However, $\text{O}_1$–$\text{O}_4$ rely primarily on limited task-specific trajectories as supervision and do not explicitly enforce alignment between generated trajectories and the intended task description $\mathbf{p}_i$.
As a result, the LLM may learn to generate coherent and physically plausible trajectories driven mainly by data supervision, while failing to respect the specified task descriptions. Such task–trajectory drift is particularly problematic in multi-task settings, where different tasks often share similar state–action distributions yet differ fundamentally in their goals.
To address this gap, we introduce two task–trajectory alignment objectives ($\text{O}_5$–$\text{O}_6$) that explicitly couple generated trajectories with their natural-language task descriptions:

\textit{Objective 5: Trajectory Explanation ($\text{O}_5$).}
$\text{O}_5$ trains the LLM to map trajectories to natural-language explanations in a supervised manner, requiring it to interpret behavioral patterns and infer the underlying task intent.
By learning to articulate what a trajectory accomplishes, the LLM develops a semantic understanding of task goals, allowing it to distinguish trajectories based on purpose rather than merely imitating plausible motion patterns.
Given a specific trajectory $\tau$ and an explanatory prompt (\emph{i.e.},  ``\texttt{Explain the following trajectory in natural language.}''), the model is trained to produce a description summarizing the task goal and the executed behavior.

\textit{Objective 6: Description-Based Rollout ($\text{O}_6$).}
$\text{O}_6$ trains the LLM to generate trajectories directly from natural-language task descriptions, enabling conditional planning across diverse tasks based on semantic understanding.
Given a task description $\mathbf{p}_i$, an initial state $s_0$, and an expected return $r$, the model is trained to produce a trajectory $\tilde{\tau}$ that fulfills $\mathbf{p}_i$.
By explicitly grounding trajectory generation in the semantics of $\mathbf{p}_i$, $\text{O}_6$ ensures that the LLM produces task-aligned behaviors—an ability that is critical in multi-task settings where different tasks may share similar initial states or final rewards yet require fundamentally different goals and behaviors.

$\text{O}_5$ and $\text{O}_6$ form a bidirectional pair that establishes a semantic feedback loop: $\text{O}_5$ maps trajectories to task descriptions ($\tau \rightarrow \mathbf{p}_i$), while $\text{O}_6$ performs the inverse mapping from task descriptions to trajectories ($\mathbf{p}_i \rightarrow \tau$).
When interleaved with $\text{O}_1$–$\text{O}_4$, these alignment objectives prevent the model from drifting toward goal-agnostic generation and ensure that augmented trajectories are not only physically plausible but also semantically consistent with their intended task descriptions.%

\subsubsection{Multi-Objective LLM Fine-Tuning.}\label{sec:finetune}

To mitigate gradient interference among the heterogeneous supervision objectives $\text{O}_1$--$\text{O}_6$, we introduce an objective-aware fine-tuning strategy utilizing LoRA~\citep{hu2022lowrank}.
Specifically, each objective $\text{O}_j$ ($j = 1, \ldots, 6$) is associated with an objective-specific LoRA adapter $\bm{\theta}_{\text{LoRA}}^{\text{O}_j}$, while the base LLM parameters $\bm{\theta}_{\text{base}}$ remain frozen (as shown in Fig.~\ref{fig:framework}).
During training, each objective updates only its corresponding adapter, effectively isolating task-specific gradients and stabilizing multi-objective optimization. 
However, isolating objectives via separate adapters does not explicitly promote knowledge sharing across objectives.
To facilitate cross-objective knowledge transfer, we introduce an additional shared LoRA adapter $\bm{\theta}_{\text{LoRA}}^{\text{sh}}$, which is updated jointly by all objectives.
Concretely, when optimizing objective $\text{O}_j$, we activate both the objective-specific adapter $\bm{\theta}_{\text{LoRA}}^{\text{O}_j}$ and the shared adapter $\bm{\theta}_{\text{LoRA}}^{\text{sh}}$, and minimize:
$\mathcal{L}_{\text{O}_j}\!\left(\theta_{\text{base}} + \theta_{\text{LoRA}}^{\text{O}_j} + \theta_{\text{LoRA}}^{\text{sh}}\right), j \in \{1, \ldots, 6\}.$
In this design, the objective-specific adapters capture specialized capabilities required by objective, while the shared adapter aggregates transferable knowledge across objectives.

\subsection{Multi-Task Policy Learning with LaMA-DT}
\label{sec:pdt}

\subsubsection{Data Preparation for LaMA-DT.}
After fine-tuning the LLM with $\text{O}_1$--$\text{O}_6$, we perform task-wise trajectory augmentation using the forward generation objective $\text{O}_3$.
For each task $\mathcal{T}_i$, we sample initial state--action pairs $(s_0,a_0)$ from the real offline dataset $\mathcal{D}_i$ and generate $N$ candidate trajectories:
$\{\tilde{\tau}^{k}\}_{k=1}^{N} \sim p_{\text{O}_3}\!\left(\tau \mid \mathbf{p}_i,(s_0,a_0)\right).$

\textbf{Augmented Trajectory Scoring and Filtering.}
Not all generated trajectories are equally reliable or task-consistent. To filter out low-quality or misaligned generations, we score each candidate trajectory $\tilde{\tau}$ using all six objectives ($\text{O}_1$–$\text{O}_6$) via likelihood-based evaluation without parameter updates, reusing the previously trained objective-specific LoRA adapters ($\bm{\theta}_{\text{LoRA}}^{\text{O}_1}$-$\bm{\theta}_{\text{LoRA}}^{\text{O}_6}$).
Concretely, for $\text{O}_1$, we measure transition consistency as
\begin{equation}
S_{\text{O}_1}(\tilde{\tau})
=\frac{1}{T}\sum_{t=0}^{T-1}
\log p_{\text{O}_1}(s_{t+1}\mid \mathbf{p}_i,s_t,a_t).
\end{equation}

For $\text{O}_2$, we sample a subtrajectory from $\tilde{\tau}$ by selecting a start index $u$ and fixing the terminal state $s_T$, and score the likelihood of reconstructing the intermediate segment as:
\begin{equation}
S_{\text{O}_2}(\tilde{\tau})
=
\frac{1}{|\tilde{\tau}_{u:T}-1|}
\sum_{t=u+1}^{T-1}
\log p_{\text{O}_2}\!\left(
s_t,a_t,r_t
\,\middle|\,
\mathcal{C}_2(t)
\right),
\end{equation}
where $\tilde{\tau}_{u:T} := \{(s_t,a_t,r_t)\}_{t=u}^{T-1}$ denotes the intermediate portion of $\tilde{\tau}$ , and the conditioning context is
$\mathcal{C}_2(t)
=
\left(\mathbf{p}_i,\,\,s_T,\,\tilde{\tau}_{u:t}\right)$.

The rollout-based objectives—forward generation ($\text{O}_3$), backward generation ($\text{O}_4$), and description-based rollout ($\text{O}_6$)—share the same autoregressive likelihood form:
\begin{equation}
S_{\text{O}_j}(\tilde{\tau})
=\frac{1}{T}\sum_{t=0}^{T-1}
\log p_{\text{O}_j}(s_{t+1},a_{t+1},r_{t+1}
\mid \mathcal{C}_j,\tilde{\tau}_{\le t}),
\end{equation}
where the conditioning context $\mathcal{C}_j$ differs across objectives:
$\text{O}_3$ conditions on the task description and initial state–action pair $(\mathbf{p}_i,s_0,a_0)$;
$\text{O}_4$ conditions on the task description and terminal state $s_T$;
and $\text{O}_6$ conditions on the task description $\mathbf{p}_i$ together with $(s_0,r)$.

For $\text{O}_5$, we evaluate alignment by scoring the task description under the trajectory-to-text model,
\begin{equation}
S_{\text{O}_5}(\tilde{\tau})
=\frac{1}{|\mathbf{p}_i|}
\sum_{\ell=1}^{|\mathbf{p}_i|}
\log p_{\text{O}_5}(p_{i,\ell}\mid \tilde{\tau},p_{i,<\ell}).
\end{equation}
All scores are length-normalized.
We aggregate the six scores with the mean
$S(\tilde{\tau})
\triangleq
\frac{1}{6}\sum_{j=1}^{6} S_{\text{O}_j}(\tilde{\tau}).$

For each task, we retain the top-$K$ candidates with the highest overall scores and add them to the augmented dataset:
$\mathcal{D}^{\text{aug}}_i
= \text{Top-}K \Big(\{\tilde{\tau}^{k}\}_{k=1}^{N}\ ;\ S(\cdot)\Big),$
and train the final policy on the combined dataset $\mathcal{D}_i \cup \mathcal{D}^{\text{aug}}_i$.

\subsubsection{Policy Learning with LaMA-DT}

\paragraph{LaMA-DT Initialization.}
After data preparation, we fine-tune LaMA-DT by repurposing the same LLM as a language-conditioned DT and attaching a policy-specific LoRA adapter $\bm{\theta}_{\text{LoRA}}^{\pi}$.
The policy adapter is initialized from the shared adapter learned during trajectory augmentation:
$\bm{\theta}_{\text{LoRA}}^{\pi} \leftarrow \theta_{\text{LoRA}}^{\text{sh}}.$
This initialization explicitly transfers shared knowledge acquired from $\text{O}_1$--$\text{O}_6$
into the policy learning stage, ensuring that the augmentation objectives directly benefit multi-task policy learning.

\paragraph{LaMA-DT Policy Learning.}
Using the prepared dataset $\mathcal{D}_i \cup \mathcal{D}^{\text{aug}}_i$, each trajectory $\tau$—real or augmented—is serialized into a sequence of triples
$\tau = \{(\hat{R}_t, s_t, a_t)\}_{t=0}^{T},$
where $\hat{R}_t = \sum_{t'=t}^{T} r_{t'}$ denotes the return-to-go.
Each trajectory is prepended with two forms of conditioning information:
(i) a provenance token $z_{\text{prov}} \in \{\texttt{real},\,\texttt{augmented}\}$ indicating whether the trajectory is real or augmented, and
(ii) the natural-language task description $\mathbf{p}_i$.
The provenance token explicitly informs the LLM of the data source, enabling LaMA-DT to perform source-aware behavior cloning.
This allows the policy to benefit from the diversity of augmented trajectories while remaining anchored to the empirical distribution of real data. %
We finally optimize LaMA-DT using Equation~\ref{eq:lama_dt_loss}.
The loss minimizes the mean squared error between ground-truth and predicted actions:
\begingroup
\small
\begin{equation}
\label{eq:lama_dt_loss}
\mathcal{L}_{\text{LaMA-DT}}
\triangleq
\mathbb{E}_{\substack{\mathcal{T}_i \sim \mathcal{T} \\
\tau \sim \mathcal{D}_i \cup \mathcal{D}_i^{\text{aug}}}}
\left[
\sum_{t=1}^{T}
\left\|
a_t - \pi\!\left(\tau_{0:t-1},\, \mathbf{p}_i,\, z_{\text{prov}}\right)
\right\|_2^{2}
\right].
\end{equation}
\endgroup

\paragraph{Testing and Generalization.}
At test time, given a task description $\mathbf{p}_i$, we set $z_{\text{prov}}=\texttt{real}$ and follow the standard DT rollout procedure.
Starting from the initial observation, the model autoregressively predicts actions conditioned on the history $(\hat{R}_{\le t}, s_{\le t}, a_{<t})$ and executes them in the environment.
Notably, LaMA-DT can be readily adapted to unseen tasks via few-shot learning: a small set of task-specific trajectories from the new task is used to quickly adapt the objective-specific LoRA adapters for data augmentation, after which scoring-and-filtering and policy learning are performed to obtain an adapted policy.
The detailed algorithm for LaMA-DT is in Alg~\ref{alg:LaMA-DT}.

\section{Experiments}
\label{sec:exp}
Our experiments are designed to answer the questions:
\textbf{(Q1)} How does LaMA-DT compare with strong offline MTRL baselines on benchmarks?
\textbf{(Q2)} In low-data regimes, does LaMA-DT retain advantages over data-hungry multi-task DT variants?
\textbf{(Q3)} Does LaMA-DT exhibit consistent performance gains across offline datasets of varying quality?
\textbf{(Q4)} How does the amount of augmented data influence performance in both standard- and low-data regimes?
\textbf{(Q5)} Are the individual components of LaMA-DT critical?
\textbf{(Q6)} Are the augmented data in LaMA-DT high-quality?%

\begin{figure}[t]
\centering
\includegraphics[width=1\textwidth]{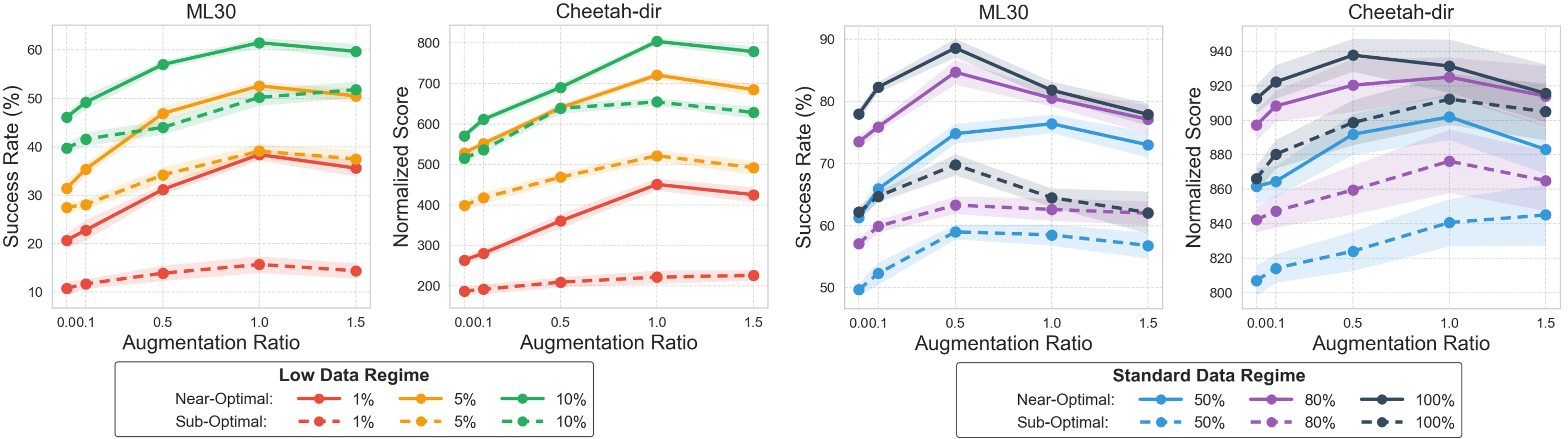}
\caption{
Performance of LaMA-DT with different augmentation ratios
$r = |\mathcal{D}^{\text{aug}}_i| / |\mathcal{D}_i|$
on Meta-World ML30 and MuJoCo Cheetah-dir, across low and standard data regimes, and under sub-optimal and near-optimal data qualities.} %
\label{fig:ratio}
\end{figure}

\paragraph{Experimental Setup.}
We evaluate LaMA-DT on two widely used offline MTRL benchmarks: MuJoCo~\citep{todorov2012mujoco} and Meta-World~\citep{yu2020meta}.
For MuJoCo locomotion, we follow prior prompt-based Decision Transformer protocols~\citep{xu2022prompting} and report results on Cheetah-dir, Cheetah-vel, and Ant-dir.
For Meta-World, we use the ML5 and ML30 task suites with shared Sawyer robot dynamics, and evaluate performance using the average \emph{success rate} across tasks.
Offline datasets are collected using SAC~\citep{haarnoja2018soft}.
We consider two dataset settings: (i) a \emph{near-optimal} setting consisting primarily of expert trajectories, and (ii) a \emph{sub-optimal} setting comprising 50\% expert and 50\% non-expert trajectories. 
For low data regimes, we uniformly subsample the per-task training data to \{1\%, 5\%, 10\%\} of the full dataset. All results are averaged over 3 seeds, and each task is evaluated for 50 episodes.

\textbf{Baselines.}
We compare LaMA-DT against representative offline MTRL baselines:
(i) \textbf{CDS}~\citep{yu2021conservative}, which applies a conservative transition-sharing strategy across tasks;
(ii) \textbf{MTBC}~\citep{hu2024harmodt}, a multi-task behavior cloning approach using network scaling and task-ID–conditioned actors;
(iii) \textbf{MTIQL}~\citep{IQL}, which extends IQL to the multi-task setting with task-conditioned actors and critics;
(iv) \textbf{MTDIFF}~\citep{he2023diffusion}, a diffusion-based generative planner for multi-task RL;
(v) \textbf{MTDT}~\citep{chen2021decision}, a multi-task DT conditioned on task IDs;
(vi) \textbf{Prompt-DT}~\citep{xu2022prompting}, which leverages trajectory prompts and return-to-go conditioning on DT; and
(vii) \textbf{HarmoDT}~\citep{hu2024harmodt}, a prompt-based DT with task-specific masking.

\begin{table}[t]
\centering
\setlength{\abovecaptionskip}{2pt}
\setlength{\belowcaptionskip}{2pt}
\caption{The average returns of LaMA-DT vs. baselines on Meta-World (Sub-Optimal) and MuJoCo (Near-Optimal) datasets under standard data regime.}
\label{tab:main_results}
\setlength{\tabcolsep}{3pt}
\renewcommand{\arraystretch}{1.0}
\scriptsize
\begin{tabular}{ll|ccccccc|c}
\toprule
\multicolumn{2}{c|}{\textbf{Task}} & \textbf{CDS} & \textbf{MTBC} & \textbf{MTIQL} & \textbf{MTDIFF} & \textbf{MTDT} & \textbf{Prompt-DT} & \textbf{HarmoDT} & \textbf{LaMA-DT} \\
\midrule
\multirow{2}{*}{\rotatebox{90}{\!\!\textbf{MW}\!\!}}
 & ML5  & 50.7{\tiny$\pm$3.3} & 46.2{\tiny$\pm$2.8} & 61.1{\tiny$\pm$4.5} & 76.0{\tiny$\pm$2.8} & 74.5{\tiny$\pm$4.4} & 76.8{\tiny$\pm$3.0} & \underline{78.9{\tiny$\pm$3.8}} & \textbf{83.4{\tiny$\pm$2.9}} \\
 & ML30 & 32.9{\tiny$\pm$2.5} & 34.5{\tiny$\pm$1.7} & 43.3{\tiny$\pm$2.2} & 57.5{\tiny$\pm$1.5} & 55.2{\tiny$\pm$2.1} & 58.7{\tiny$\pm$1.3} & \underline{63.4{\tiny$\pm$1.6}} & \textbf{69.8{\tiny$\pm$1.7}} \\
\midrule
\multirow{3}{*}{\rotatebox{90}{\!\!\textbf{MuJoCo}\!\!}}
 & Cheetah-vel & -134.5{\tiny$\pm$15.8} & -148.4{\tiny$\pm$11.2} & -142.0{\tiny$\pm$14.4} & -186.2{\tiny$\pm$8.2} & -170.1{\tiny$\pm$17.9} & -127.6{\tiny$\pm$10.9} & \underline{-86.6{\tiny$\pm$7.7}} & \textbf{-73.5{\tiny$\pm$12.5}} \\
 & Cheetah-dir & 428.9{\tiny$\pm$17.5} & 429.7{\tiny$\pm$6.6} & 454.1{\tiny$\pm$21.3} & 536.8{\tiny$\pm$10.7} & 678.5{\tiny$\pm$24.3} & 705.6{\tiny$\pm$8.3} & \underline{926.3{\tiny$\pm$10.4}} & \textbf{937.8{\tiny$\pm$9.6}} \\
 & Ant-dir     & 131.0{\tiny$\pm$10.6} & 153.8{\tiny$\pm$15.9} & 120.6{\tiny$\pm$9.9} & 248.0{\tiny$\pm$16.2} & 235.7{\tiny$\pm$15.4} & 302.5{\tiny$\pm$17.1} & \underline{315.6{\tiny$\pm$5.5}} & \textbf{328.5{\tiny$\pm$8.4}} \\
\bottomrule
\end{tabular}
\end{table}

\paragraph{Main Results (Q1).}
Table~\ref{tab:main_results} summarizes average performance on Meta-World (ML5/ML30) and MuJoCo locomotion tasks (Cheetah-vel, Cheetah-dir, Ant-dir) under standard data regimes.
LaMA-DT consistently outperforms all baselines across benchmarks, achieving an average improvement of 5.5\% on Meta-World and 6.8\% on MuJoCo relative to the strongest baseline.
On Meta-World, gains are most pronounced on ML30, a highly heterogeneous task suite prone to negative transfer, where LaMA-DT improves over the best baseline by 6.4\%.
On MuJoCo, LaMA-DT performs strongly on both direction- and velocity-conditioned control, achieving the largest improvements on Cheetah-vel (15.1\%) and Ant-dir (4.1\%), while remaining competitive on Cheetah-dir.
These results indicate that even under standard data regimes, the proposed data augmentation strategy in LaMA-DT continues to provide consistent performance gains.

\paragraph{Performance in Low-Data Regimes (Q2).}
In Table~\ref{tab:lowdata}, across all benchmarks and data fractions, LaMA-DT consistently outperforms DT-based baselines, demonstrating substantial advantages in low-data regimes.  
It achieves consistent gains on both ML30 and Cheetah-dir across all data levels (1\%, 5\%, and 10\%), with performance improving steadily as more data become available.  
For example, on ML30 with suboptimal offline data, LaMA-DT improves from 15.7 at 1\% data to 51.8 at 10\%, significantly outperforming Prompt-DT (by 22.7\% at 1\% and 40.8\% at 10\%) and HarmoDT (by 29.8\% at 1\% and 21.6\% at 10\%), indicating superior data efficiency.  
A similar trend is observed on Cheetah-dir: under near-optimal data, it outperforms HarmoDT by 47.3\% and Prompt-DT by 58.5\% on average across the three data fractions.  
Notably, even in the extreme 1\% regime, LaMA-DT maintains a clear margin over all baselines, highlighting its robustness under severe data scarcity.

\paragraph{Robustness to Data Quality (Q3).}
Table~\ref{tab:lowdata} shows that LaMA-DT consistently maintains a clear performance advantage over competing methods under both \emph{sub-optimal} and \emph{near-optimal} datasets.
Notably, LaMA-DT substantially outperforms DT-based baselines in the \emph{sub-optimal} setting, demonstrating strong robustness to imperfect offline data.
As data quality improves, LaMA-DT exhibits larger absolute performance gains than competing approaches:
on Meta-World ML30 at the 5\% data level, performance increases from 39.1 (sub-optimal) to 52.6 (near-optimal), a gain that is significantly larger than those achieved by Prompt-DT or HarmoDT.
These results indicate that LaMA-DT can handle low-quality data effectively, and scale favorably with improved data quality, suggesting the augmented data is of high quality and successfully mitigates the data imperfection issues.

\begin{table}[t]
\caption{Performance under low-data regimes (1\%, 5\%, and 10\%) across different data qualities (sub-optimal vs.\ near-optimal). %
}

\label{tab:lowdata}
\setlength{\tabcolsep}{2pt}
\renewcommand{\arraystretch}{1.1}
\resizebox{\textwidth}{!}{
\begin{tabular}{l|ccc|ccc|ccc|ccc}
\toprule
\multirow{3}{*}{\textbf{Method}}
& \multicolumn{6}{c|}{\textbf{ML30}}
& \multicolumn{6}{c}{\textbf{Cheetah-dir}} \\
\cmidrule(lr){2-7}\cmidrule(lr){8-13}
& \multicolumn{3}{c|}{\textbf{Sub-Optimal}}
& \multicolumn{3}{c|}{\textbf{Near-Optimal}}
& \multicolumn{3}{c|}{\textbf{Sub-Optimal}}
& \multicolumn{3}{c}{\textbf{Near-Optimal}} \\
\cmidrule(lr){2-4}\cmidrule(lr){5-7}\cmidrule(lr){8-10}\cmidrule(lr){11-13}
& \textbf{1\%} & \textbf{5\%} & \textbf{10\%}
& \textbf{1\%} & \textbf{5\%} & \textbf{10\%}
& \textbf{1\%} & \textbf{5\%} & \textbf{10\%}
& \textbf{1\%} & \textbf{5\%} & \textbf{10\%} \\
\midrule
MTDT
& 10.9$\pm$3.0 & 19.2$\pm$2.4 & 25.6$\pm$2.3
& 12.5$\pm$2.0 & 24.1$\pm$2.3 & 35.6$\pm$1.9
& 113.4$\pm$20.2 & 274.5$\pm$18.0 & 386.3$\pm$21.9
& 145.2$\pm$23.1 & 320.5$\pm$19.6 & 435.6$\pm$20.4 \\
Prompt-DT
& 12.8$\pm$1.4 & 24.3$\pm$0.9 & 36.8$\pm$1.1
& 15.3$\pm$1.5 & 28.5$\pm$0.7 & 41.2$\pm$0.8
& 191.3$\pm$17.1 & 378.9$\pm$18.3 & 507.4$\pm$15.3
& 256.3$\pm$16.2 & 474.0$\pm$12.8 & 544.7$\pm$15.7 \\
HarmoDT
& 12.1$\pm$1.6 & 27.0$\pm$1.4 & 42.6$\pm$1.5
& 22.4$\pm$1.1 & 36.1$\pm$0.9 & 46.8$\pm$0.9
& 185.2$\pm$12.3 & 452.2$\pm$17.5 & 526.8$\pm$13.9
& 260.7$\pm$11.2 & 540.3$\pm$9.8 & 592.0$\pm$10.3 \\
\midrule
\textbf{LaMA-DT}
& \textbf{15.7$\pm$1.7} & \textbf{39.1$\pm$1.5} & \textbf{51.8$\pm$1.7}
& \textbf{38.4$\pm$1.2} & \textbf{52.6$\pm$0.9} & \textbf{61.5$\pm$1.0}
& \textbf{225.9$\pm$12.7} & \textbf{521.5$\pm$14.0} & \textbf{654.4$\pm$9.2}
& \textbf{450.6$\pm$15.1} & \textbf{720.8$\pm$9.4} & \textbf{803.9$\pm$8.9} \\
\bottomrule
\end{tabular}
}
\end{table}

\paragraph{Effect of Augmentation Ratio (Q4).}
We examine how the augmentation ratio $r = |\mathcal{D}^{\text{aug}}_i| / |\mathcal{D}_i|$ affects LaMA-DT performance under different data quantities (low-data: 1\%, 5\%, 10\%; standard-data: 50\%, 80\%, 100\%) and data qualities (sub-optimal vs.\ near-optimal).
In low-data regimes (Fig.~\ref{fig:ratio}(a,c)), increasing $r$ substantially improves performance, confirming that trajectory augmentation effectively mitigates data scarcity.
However, the gains are non-monotonic: overly large ratios (\emph{e.g.}, $r=1.5$) often lead to saturation or degradation, suggesting that excessive augmented data may introduce distribution shift and error accumulation.
In standard-data regimes (Fig.~\ref{fig:ratio}(b,d)), the optimal ratio shifts to smaller values (typically $r \approx 0.5$), as abundant real data already provides strong supervision.
Across all subfigures, LaMA-DT trained on near-optimal data (solid lines) exhibit more stable improvements than those trained on sub-optimal data (dashed lines), indicating that augmentation cannot fully compensate for low-quality data and may amplify suboptimal patterns when overused.

\begin{table}[!ht]
\centering
\setlength{\abovecaptionskip}{2pt}
\setlength{\belowcaptionskip}{2pt}
\captionsetup{singlelinecheck=false, justification=raggedright}
\caption{Ablation studies on LaMA-DT.}
\label{tab:ablate_tasks}
\setlength{\tabcolsep}{3pt}
\renewcommand{\arraystretch}{1.05}
\scriptsize

\resizebox{\linewidth}{!}{%
\begin{tabular}{l|cccccc}
\toprule
\textbf{Task}
& w/o $\text{O}_1$ & w/o $\text{O}_2$ & w/o $\text{O}_3$ & w/o $\text{O}_4$ & w/o $\text{O}_5$ & w/o $\text{O}_6$ \\
\midrule
ML30        & 61.7$\pm$1.3   & 60.5$\pm$2.1   & 56.4$\pm$2.8   & 66.3$\pm$1.6   & 63.8$\pm$2.0   & 65.0$\pm$2.3 \\
Cheetah-dir & 842.3$\pm$14.3 & 862.4$\pm$10.2 & 765.9$\pm$11.7 & 910.2$\pm$16.8 & 902.6$\pm$15.4 & 922.5$\pm$26.3 \\
Ant-dir     & 259.0$\pm$7.0  & 281.3$\pm$8.5  & 247.3$\pm$9.6  & 310.8$\pm$12.3 & 313.9$\pm$10.4 & 322.4$\pm$16.2 \\
\bottomrule
\end{tabular}%
}

\vspace{3pt}

\resizebox{\linewidth}{!}{%
\begin{tabular}{l|cc|cc|c}
\toprule
\textbf{Task}
& w/o $\text{O}_5$\&$\text{O}_6$ & w/o $\text{O}_2$-$\text{O}_4$
& w/o $\theta_{\text{LoRA}}^{\text{sh}}$ & w/o Filtering
& \textbf{LaMA-DT} \\
\midrule
ML30        & 57.8$\pm$2.0   & 50.2$\pm$2.7   & 63.1$\pm$1.4 & 62.1$\pm$3.4   & \textbf{69.8$\pm$1.7} \\
Cheetah-dir & 835.1$\pm$13.5 & 636.9$\pm$22.1 & 878.0$\pm$8.4 & 883.1$\pm$15.9 & \textbf{937.8$\pm$9.6} \\
Ant-dir     & 274.7$\pm$8.9  & 216.6$\pm$8.7  & 313.2$\pm$9.8 & 285.8$\pm$12.6 & \textbf{328.5$\pm$8.4} \\
\bottomrule
\end{tabular}%
}
\end{table}

\paragraph{Ablation Studies (Q5).} As shown in Table~\ref{tab:ablate_tasks}, removing any component 

results in performance degradation. 
Among single-objective ablations, 
$\text{O}_3$ contributes the most, confirming the importance of long-horizon 
trajectory structure.
More importantly, group ablations reveal strong synergy: jointly removing the Sequential Decision Learning objectives (w/o $\text{O}_2$, $\text{O}_3$, and $\text{O}_4$) results in the largest performance drop across all benchmarks (50.2 vs.\ 69.8 on ML30; 636.9 vs.\ 937.8 on Cheetah-dir), substantially greater than any single-objective ablation. This highlights the effectiveness of trajectory-level supervision as a unified group.
Similarly, removing both task-level objectives (w/o $\text{O}_5$\&$\text{O}_6$) causes a sharper drop than either alone, showing their complementary role in mitigating task drift. %

\begin{table}[t]
\centering
\setlength{\abovecaptionskip}{2pt}
\setlength{\belowcaptionskip}{2pt}
\caption{
Quality of augmented data on Cheetah-dir. We compare against real expert data (reference) and MTDIFF~\citep{he2023diffusion} as a  generator baseline.
Lower values are better for Dyn.\ MSE, State KL, and L2 Dist., while higher values are better for Valid Coverage. Criteria details are in Appendix~\ref{app:aug_quality_metrics}.
}
\label{tab:aug_quality}
\setlength{\tabcolsep}{3.5pt}
\renewcommand{\arraystretch}{1.1}
\small
\begin{tabular}{l|cccc}
\toprule
\textbf{Source}
& \textbf{Dyn.\ MSE}$\downarrow$
& \textbf{State KL vs.\ real}$\downarrow$
& \textbf{L2 Dist.\ from Dataset}
& \textbf{Valid Coverage}$\uparrow$ \\
\midrule
Real expert (reference)         & 0.0000          & 0.00          & ---     & 100\%          \\
MTDIFF~\citep{he2023diffusion}   & 0.0107          & 0.51          & 0.4552  & 102\%          \\
LaMA-DT (unfiltered)               & 0.0089          & 0.42          & 0.4831  & 99\%           \\
\textbf{LaMA-DT (filtered)}        & \textbf{0.0031} & \textbf{0.18} & \textbf{0.4376}  & \textbf{109\%} \\
\bottomrule
\end{tabular}
\end{table}

\paragraph{Augmentation Quality (Q6).}

\begin{figure}[!ht]
    \centering
    \includegraphics[width=0.7\linewidth]{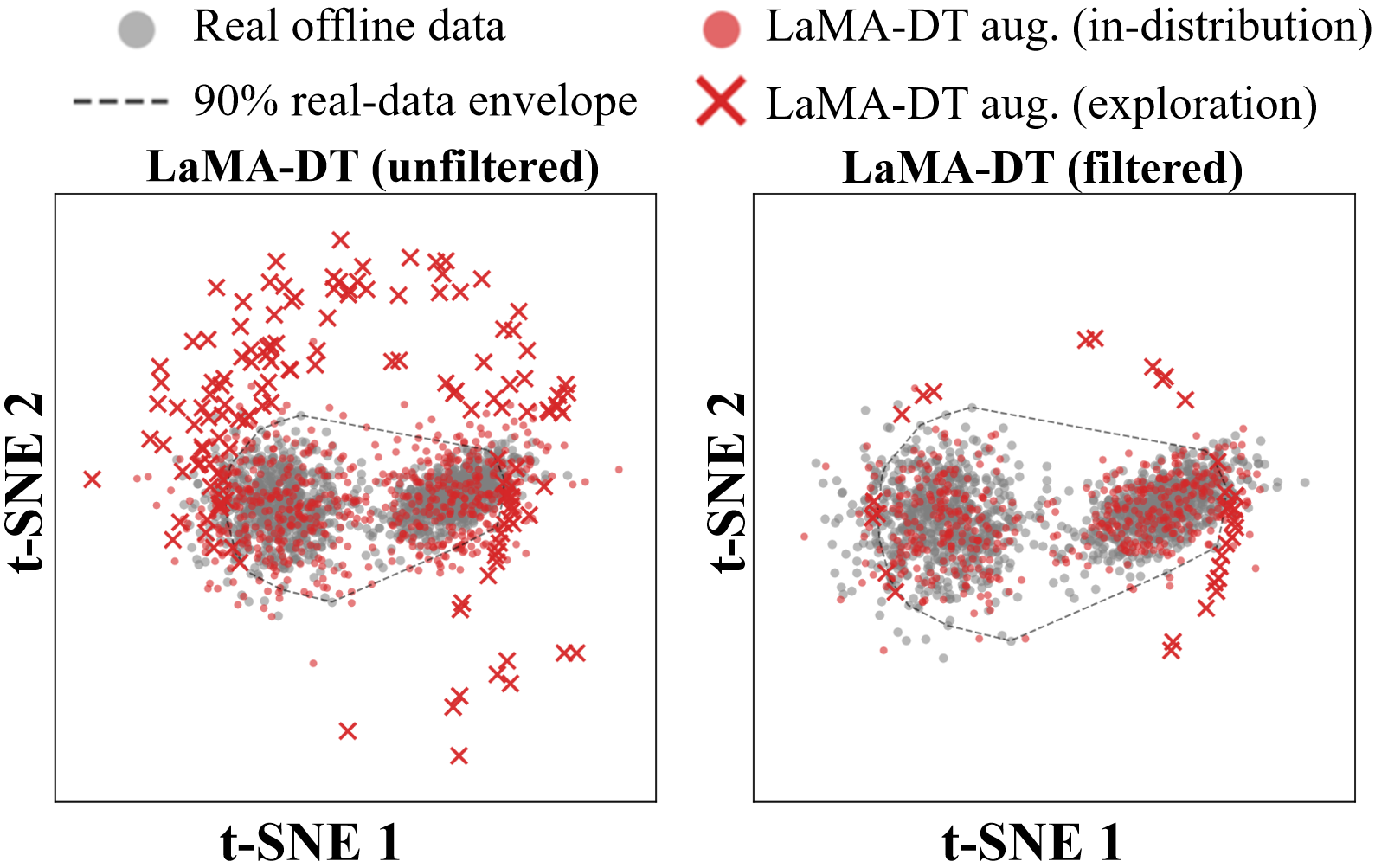}

    \caption{t-SNE projection of Cheetah-dir states:
  \textbf{unfiltered} (left) vs.\ \textbf{filtered} (right).
  Gray dots are real states from offline data, red dots are augmented states overlapping the real manifold (in-distribution), red crosses are augmented states lying outside the dataset for exploration,
  and the dashed contour is the $90\%$ real-data envelope.}
    \label{fig:tsne}
\end{figure}

We assess augmentation quality on Cheetah-dir using four criteria reported in Tab.~\ref{tab:aug_quality}: Dyn.\ MSE for dynamics fidelity; State KL for distributional alignment with real data; L2 Dist.\ (to real states) for training-set proximity; and Valid Coverage, which measures the fraction of physically valid generated states relative to real states (values above $100\%$ indicate out-of-distribution yet effective exploratory states validated by the environment dynamics).
Against MTDIFF~\citep{he2023diffusion}, our filtered augmentations achieve $\sim$$3.5\times$ lower Dyn.\ MSE and $\sim$$2.8\times$ lower State KL, indicating stronger physical faithfulness and tighter distributional alignment. Crucially, the filter in LaMA-DT does not simply prune novel states: comparing unfiltered to filtered, it lowers Dyn.\ MSE ($0.0089 \!\rightarrow\! 0.0031$), halves State KL ($0.42 \!\rightarrow\! 0.18$), \emph{and} raises Valid Coverage from $99\%$ to $109\%$, indicating selectively removing off-manifold rollouts while retaining $9\%$ of dynamics-validated novel states beyond the offline data. %
Fig.~\ref{fig:tsne} corroborates this: the unfiltered panel scatters exploration crosses far outside the $90\%$ real-data envelope, while the filtered panel keeps the in-distribution cloud tightly overlapping the real clusters and an exploration set that hugs the envelope from just outside. These quality gains translate to clear downstream policy improvements over the unfiltered variant ($+54.7$ on Cheetah-dir) shown in Table~\ref{tab:ablate_tasks}.

\FloatBarrier
\section{Conclusion}
We presented LaMA-DT, an LLM-based offline MTRL framework designed to address data scarcity and cross-task generalization. LaMA-DT leverages specialized fine-tuning objectives to enable a single LLM to generate high-fidelity, task-aligned trajectories for augmentation. The same LLM is then repurposed as a DT and fine-tuned on both augmented and real trajectories. Experiments demonstrate that LaMA-DT consistently outperforms state-of-the-art offline MTRL baselines, highlighting the effectiveness of LLM-driven trajectory augmentation and policy learning.

\subsection*{Reproducibility statement}

We provide pseudocode for the three stages of LaMA-DT in Appendix~B,
benchmark and dataset details in Appendix~C, baseline descriptions in
Appendix~D, and implementation settings, hyperparameters, and evaluation
protocols in Appendix~E. An anonymized code repository is available at
\url{https://anonymous.4open.science/r/o33zlamadt-6587}.

\bibliographystyle{iclr2027_conference}
\bibliography{example_paper}

\newpage
\appendix

\section{Related Work}
\label{sec:related_work}

\noindent\textbf{Offline RL.}
Offline reinforcement learning learns policies from fixed datasets without additional environment interaction~\citep{levine2020offline,batchRL,batchRL2}, and has shown promise in robotics~\citep{Kalashnikov2018,Rafailov2020}, NLP~\citep{Jaques2019}, and healthcare~\citep{Wang2018}.
A central challenge is \emph{distribution shift} between the learned policy and the behavior policy, which can lead to erroneous value backups and unstable learning~\citep{Fujimoto2018,levine2020offline}.
To address this, prior work constrains the policy to stay close to the behavior distribution~\citep{Kumar2019,Fujimoto2018,Wu2019,Jaques2019,Peng2019,Siegel2020} or regularizes the learned value/Q functions~\citep{CQL,Kostrikov2021}, while model-based approaches build conservative dynamics/reward models to reduce out-of-distribution errors~\citep{MOReL,MOPO}.
In addition, Decision Transformer~\citep{chen2021decision} casts offline RL as conditional sequence modeling.

\noindent\textbf{Multi-task RL and offline MTRL.}
Multi-task RL studies learning a shared policy across tasks in RL settings.
A key obstacle is negative transfer caused by conflicting gradients across tasks, motivating gradient-surgery or conflict-aware optimization methods~\citep{yu2020gradient,chen2020just,liu2021conflict}.
Multi-task offline RL aims to learn a shared policy from offline datasets collected from a set of tasks.
A common formulation is to train a task-conditioned policy under a unified objective, but such shared training often suffers from negative transfer caused by conflicting gradients across tasks and heterogeneous data sources.
To mitigate this issue, prior work has explored selective sharing and conflict-aware optimization mechanisms.
HarmoDT~\citep{hu2024harmodt} addresses this challenge from a parameter-selection perspective by learning task-specific binary masks over shared parameters, enabling conflict-aware parameter sharing in multi-task offline Decision Transformer training.
Complementary to gradient conflict handling, data sharing methods selectively reuse transitions across tasks while remaining conservative under distribution shift, such as CDS~\citep{yu2021conservative} and related value-based sharing approaches~\citep{Eysenbach2020,MT-Opt,UDS}.
More recently, M3DT~\citep{kong2025m3dt} introduces a Mixture-of-Experts Decision Transformer to scale offline MTRL to massive task suites by decoupling task load across expert subsets. However, M3DT targets \emph{task scalability under full per-task datasets}, an axis orthogonal to our focus on \emph{per-task data scarcity}: LaMA-DT addresses scarcity through language-guided trajectory augmentation rather than through architectural expansion.
In contrast to these baselines, LaMA-DT aligns text and behavior, and then uses the resulting objectives to score and filter generated trajectories before decision-transformer policy learning, leading to more robust multi-task offline RL.

\noindent\textbf{Offline Meta-RL.}
Offline meta-RL aims to learn policies that generalize to unseen tasks from fixed offline datasets, without any online interaction. A large body of prior work follows the temporal-difference (TD) paradigm, using either optimization-based meta-learning to find fast-adaptable initializations~\citep{MAML,Mitchell2020OfflineML,xu2018meta_rl}.
However, TD-based offline meta-RL can be prone to instability under off-policy bootstrapping~\citep{levine2020offline}, motivating alternative approaches that avoid value backups.
Along this line, decision-transformer-style methods treat offline meta-RL as conditional sequence modeling over trajectories~\citep{chen2021decision}.
Prompt-based approaches such as Prompt-DT~\citep{xu2022prompting} and DPDT~\citep{zheng2024dpdt} adapt to new tasks using soft or hand-crafted prompts, while GDT~\citep{Furuta2021GeneralizedDT} leverages hindsight rewards as an adaptation signal. Meta-DT~\citep{wang2024metadt} further introduces a meta-policy to select informative trajectories for adaptation.
Two contemporaneous approaches further inject language signals into this paradigm: LPDT~\citep{yang2025lpdt} initializes Prompt-DT with a pre-trained language model and adds prompt regularization to improve few-shot adaptation, while T2DA~\citep{zhang2025t2da} learns CLIP-style contrastive alignment between task descriptions and trajectory embeddings to enable zero-shot transfer. Both methods leverage language to ground task semantics, but they differ from LaMA-DT in two important ways: (i) neither performs \emph{trajectory-level data augmentation}---LPDT relies on LM initialization while T2DA relies on contrastive alignment---and (ii) both are evaluated primarily under the meta-RL zero/few-shot transfer protocol, whereas LaMA-DT addresses the multi-task offline setting under varying per-task data fractions. We accordingly compare against multi-task offline RL baselines that share our evaluation protocol, and view direct comparison with LPDT and T2DA on their meta-RL setups as a complementary direction.
Our work builds on the sequence-modeling perspective, but explicitly leverages natural language task descriptions to ground task semantics in behavioral data, also enabling generalization ability in offline meta-RL settings.

\noindent\textbf{LLM-based offline RL.}
A complementary line of work directly leverages pre-trained language models for offline RL. LaMo~\citep{shi2024lamo} initializes the DT backbone with a frozen pre-trained language model and fine-tunes via LoRA on a small fraction of parameters, demonstrating substantial gains in \emph{single-task} offline RL---especially under sparse rewards and limited data. LaMA-DT extends this direction along two axes: (i) we operate in the multi-task offline setting rather than per-task training, requiring explicit mechanisms to disambiguate similar state-action distributions across tasks; and (ii) we reuse the same LM not only as the policy backbone but also as a \emph{trajectory generator} that produces task-aligned augmented data via six supervised objectives, addressing the data scarcity bottleneck that initialization alone cannot resolve.

\section{Detailed Training Algorithm}
\label{app:detailed_algorithm}

We give the full procedure of LaMA-DT in two complementary forms.
Algorithm~\ref{alg:LaMA-DT} provides a compact high-level overview of the three stages, and Algorithms~\ref{alg:stage1}--\ref{alg:stage3} restate each stage with the full set of algorithmic details (sampling, adapter activation, loss instantiation), following the same notation as Sec.~\ref{sec:augmentation}--Sec.~\ref{sec:pdt}.
We use $\theta_{\text{base}}$ for the frozen LLM weights, $\theta_{\text{LoRA}}^{\text{O}_j}$ for the objective-specific LoRA adapter of objective $\text{O}_j$, $\theta_{\text{LoRA}}^{\text{sh}}$ for the shared LoRA adapter, and $\theta_{\text{LoRA}}^{\pi}$ for the policy LoRA used at Stage~III.
We write $f_{\bm{\theta}}(\cdot)$ for a forward pass with the active LoRA set $\bm{\theta}$ added on top of $\theta_{\text{base}}$.

\begin{algorithm}[H]
\caption{LaMA-DT Training (Compact Overview)}
\label{alg:LaMA-DT}
\begin{algorithmic}[1]
\STATE \textbf{Input:} $\{(\mathbf{p}_i,\mathcal{D}_i)\}_{i=1}^{M}$; frozen base LLM params $\boldsymbol{\theta}_{\text{base}}$.
\STATE \textbf{Output:} Unified multi-task policy $\pi_{\boldsymbol{\theta}}$.

\STATE \textit{// Stage I: Multi-objective LLM fine-tuning ($\text{O}_1$--$\text{O}_6$)}
\STATE Initialize objective adapters $\{\boldsymbol{\theta}_{\text{LoRA}}^{\text{O}_j}\}_{j=1}^{6}$ and shared adapter $\boldsymbol{\theta}_{\text{LoRA}}^{\text{sh}}$; freeze $\boldsymbol{\theta}_{\text{base}}$.
\WHILE{not converged}
    \STATE Sample task $i$ and a batch $\mathcal{B}\subset\mathcal{D}_i$; sample $\text{O}_j$.
    \STATE Update $\boldsymbol{\theta}_{\text{LoRA}}^{\text{O}_j}$ and $\boldsymbol{\theta}_{\text{LoRA}}^{\text{sh}}$ with the loss of $\text{O}_j$.
\ENDWHILE

\STATE \textit{// Stage II: Data preparation ($\text{O}_3$ generation + $\text{O}_1$--$\text{O}_6$ scoring)}
\FOR{each task $i=1$ to $M$}
    \STATE Initialize $\mathcal{D}^{\text{aug}}_i\leftarrow\emptyset$.
    \FOR{each sampled $(s_0,a_0)\sim\mathcal{D}_i$}
        \STATE Generate $\{\tilde{\tau}^{k}\}_{k=1}^{N}$ using $\text{O}_3$.
        \STATE Score each $\tilde{\tau}^{k}$ by $S(\tilde{\tau}^{k})=\frac{1}{6}\sum_{j=1}^{6}S_{\text{O}_j}(\tilde{\tau}^{k})$.
        \STATE Add the top-$K$ candidates $(\tilde{\tau},\mathbf{p}_i,z_{\text{prov}}=\texttt{augmented})$ to $\mathcal{D}^{\text{aug}}_i$.
    \ENDFOR
\ENDFOR

\STATE \textit{// Stage III: Policy learning (LaMA-DT)}
\STATE Initialize policy adapter $\boldsymbol{\theta}_{\text{LoRA}}^{\pi}\leftarrow \boldsymbol{\theta}_{\text{LoRA}}^{\text{sh}}$; freeze $\boldsymbol{\theta}_{\text{base}}$.
\STATE Train LaMA-DT on $\{\mathcal{D}_i\cup\mathcal{D}^{\text{aug}}_i\}$ using Eq.~\ref{eq:lama_dt_loss}.
\STATE \textbf{return} $\pi_{\boldsymbol{\theta}}$.
\end{algorithmic}
\end{algorithm}

\paragraph{Trajectory serialization.}
For each task $\mathcal{T}_i$, every trajectory $\tau = \{(s_t, a_t, r_t)\}_{t=0}^{T}$ is mapped into the LLM hidden space via three numeric MLPs (one each for $s$, $a$, the per-step return-to-go $\hat{R}_t = \sum_{t' \ge t} r_{t'}$). The natural-language task description $\mathbf{p}_i$ is tokenized and embedded by the frozen LLM tokenizer; for Stage~III an extra provenance marker $z_{\text{prov}} \in \{\texttt{real}, \texttt{augmented}\}$ is appended to the text prefix. The full input is
\[
[\,\texttt{Emb}(\mathbf{p}_i,\,z_{\text{prov}})\;\Vert\;\hat{R}_0,\,s_0,\,a_0,\;\hat{R}_1,\,s_1,\,a_1,\;\ldots\;]
\]
and is processed by $f_{\bm{\theta}}$ as a single sequence; the numeric heads read out the corresponding per-step continuous predictions.

\subsection{Stage~I: Multi-Objective LoRA Fine-Tuning}
\begin{algorithm}[H]
\caption{Stage~I: Multi-Objective LoRA Fine-Tuning}
\label{alg:stage1}
\begin{algorithmic}[1]
\STATE \textbf{Input:} task pool $\{(\mathbf{p}_i, \mathcal{D}_i)\}_{i=1}^{M}$; frozen $\theta_{\text{base}}$; iteration budget $T_{\text{I}}$; window length $L$.
\STATE Initialize objective adapters $\{\theta_{\text{LoRA}}^{\text{O}_j}\}_{j=1}^{6}$ and shared adapter $\theta_{\text{LoRA}}^{\text{sh}}$.
\FOR{step $= 1, \ldots, T_{\text{I}}$}
    \STATE Sample task $i \sim \mathrm{Uniform}(\{1,\ldots,M\})$ and trajectory $\tau \sim \mathcal{D}_i$.
    \STATE Sample window of length $L$ from $\tau$ and an objective $j \sim \mathrm{Uniform}(\{1,\ldots,6\})$.
    \STATE Construct $(x_j, y_j)$ from $(\mathbf{p}_i, \tau_{\text{window}})$ as in Sec.~\ref{sec:Trajectory Generation}--\ref{sec:alignment}.
    \STATE Activate the active set $\bm{\theta}_j = \theta_{\text{base}} + \theta_{\text{LoRA}}^{\text{O}_j} + \theta_{\text{LoRA}}^{\text{sh}}$.
    \STATE Compute the loss $\mathcal{L}_{\text{O}_j}\!\left(f_{\bm{\theta}_j}(x_j),\,y_j\right)$ \hfill (MSE for $j\!\in\!\{1,2,3,4,6\}$; CE for $j\!=\!5$).
    \STATE Update only $\theta_{\text{LoRA}}^{\text{O}_j}$ and $\theta_{\text{LoRA}}^{\text{sh}}$ by AdamW; $\theta_{\text{base}}$ stays frozen.
\ENDFOR
\STATE \textbf{Output:} adapters $\{\theta_{\text{LoRA}}^{\text{O}_j}\}_{j=1}^{6}$ and shared adapter $\theta_{\text{LoRA}}^{\text{sh}}$.
\end{algorithmic}
\end{algorithm}

\subsection{Stage~II: Augmented Trajectory Generation, Scoring, and Filtering}
\begin{algorithm}[H]
\caption{Stage~II: Augmented Trajectory Generation, Scoring, and Filtering}
\label{alg:stage2}
\begin{algorithmic}[1]
\STATE \textbf{Input:} adapters from Stage~I; $\{\mathcal{D}_i\}_{i=1}^M$; \#candidates $N$, top-$K$, sampling temperature $\mathcal{T}$.
\STATE Initialize $\mathcal{D}_i^{\text{aug}} \leftarrow \emptyset$ for all $i$.
\FOR{each task $i = 1, \ldots, M$}
    \FOR{each sampled $(s_0, a_0) \in \mathcal{D}_i$}
        \STATE Activate $\theta_{\text{base}} + \theta_{\text{LoRA}}^{\text{O}_3} + \theta_{\text{LoRA}}^{\text{sh}}$ and autoregressively roll out $N$ candidate trajectories at temperature $\mathcal{T}$:
        \par\hspace{2.6em}$\{\tilde{\tau}^{k}\}_{k=1}^{N} \sim p_{\text{O}_3}\!\left(\tau \mid \mathbf{p}_i,\,(s_0, a_0)\right)$.
        \FOR{$k = 1, \ldots, N$}
            \FOR{$j = 1, \ldots, 6$}
                \STATE Activate $\theta_{\text{base}} + \theta_{\text{LoRA}}^{\text{O}_j}$ and compute the length-normalized teacher-forcing log-likelihood $S_{\text{O}_j}(\tilde{\tau}^k)$ from Sec.~\ref{sec:pdt}.
            \ENDFOR
            \STATE Aggregate score $S(\tilde{\tau}^k) \triangleq \tfrac{1}{6}\sum_{j=1}^{6} S_{\text{O}_j}(\tilde{\tau}^k)$.
        \ENDFOR
        \STATE Add the top-$K$ candidates by $S$, each tagged with $z_{\text{prov}}=\texttt{augmented}$, to $\mathcal{D}_i^{\text{aug}}$.
    \ENDFOR
\ENDFOR
\STATE \textbf{Output:} augmented per-task datasets $\{\mathcal{D}_i^{\text{aug}}\}_{i=1}^M$.
\end{algorithmic}
\end{algorithm}

\paragraph{Stage~II scoring as a budget-cheap reranker.}
Generation calls $\theta_{\text{LoRA}}^{\text{O}_3}$ once per candidate; scoring then loops over the six objective adapters, but each scoring pass is forward-only, length-normalized, and shares the same input encoding. Because filtering keeps top-$K$ candidates per $(s_0, a_0)$ rather than per task, the augmentation budget grows linearly with the number of seed pairs and is independent of $M$.

\subsection{Stage~III: Multi-Task Policy Learning}
\begin{algorithm}[H]
\caption{Stage~III: Multi-Task Policy Learning}
\label{alg:stage3}
\begin{algorithmic}[1]
\STATE \textbf{Input:} $\{\mathcal{D}_i \cup \mathcal{D}_i^{\text{aug}}\}_{i=1}^M$; shared adapter $\theta_{\text{LoRA}}^{\text{sh}}$ from Stage~I; iteration budget $T_{\text{III}}$.
\STATE Initialize the policy adapter from the shared adapter: $\theta_{\text{LoRA}}^{\pi} \leftarrow \theta_{\text{LoRA}}^{\text{sh}}$; freeze $\theta_{\text{base}}$.
\FOR{step $= 1, \ldots, T_{\text{III}}$}
    \STATE Sample task $i$ (uniform over tasks; optional balanced re-weighting); sample $\tau \sim \mathcal{D}_i \cup \mathcal{D}_i^{\text{aug}}$ with augmentation ratio $r$.
    \STATE Sample a window of length $L$ from $\tau$ and form the DT-style input with $\mathbf{p}_i$ and $z_{\text{prov}}\in\{\texttt{real},\texttt{augmented}\}$.
    \STATE Predict $\hat{a}_t = f_{\theta_{\text{base}}+\theta_{\text{LoRA}}^{\pi}}(\tau_{0:t-1},\mathbf{p}_i,z_{\text{prov}})$ for every action position.
    \STATE Update $\theta_{\text{LoRA}}^{\pi}$ (and the numeric heads) by AdamW on $\mathcal{L}_{\text{LaMA-DT}}$ in Eq.~\ref{eq:lama_dt_loss}; $\theta_{\text{base}}$ stays frozen.
\ENDFOR
\STATE \textbf{Deployment:} given $\mathbf{p}_i$, set $z_{\text{prov}} = \texttt{real}$ and roll out autoregressively; update $\hat{R}_{t+1} = \hat{R}_t - r_t$.
\STATE \textbf{Output:} unified multi-task policy $\pi_{\theta_{\text{base}}+\theta_{\text{LoRA}}^{\pi}}$.
\end{algorithmic}
\end{algorithm}

\subsection{Per-Objective Loss Instantiation}
\label{app:loss_instantiation}
Stage~I uses a Gaussian log-likelihood (equivalently, length-normalized MSE) for the five numeric objectives and a token-level cross-entropy for the text-output objective $\text{O}_5$:
\begin{align}
\mathcal{L}_{\text{O}_j}\!\left(\bm{\theta}_j\right)
& = \frac{1}{|y_j|} \sum_{t=1}^{|y_j|}
       \big\| y_{j,t} - \mu_{j,t}(x_j;\bm{\theta}_j) \big\|_2^{2}
\qquad j \in \{1, 2, 3, 4, 6\}, \\[2pt]
\mathcal{L}_{\text{O}_5}\!\left(\bm{\theta}_5\right)
& = -\frac{1}{|y_5|}\sum_{t=1}^{|y_5|}
       \log p_{\bm{\theta}_5}\!\left(y_{5,t} \mid x_5,\, y_{5,<t}\right).
\end{align}
The same $(x_j, y_j)$ construction is reused at scoring time (Algorithm~\ref{alg:stage2}, line 9), so the resulting $S_{\text{O}_j}(\tilde{\tau})$ are interpretable as conditional log-likelihoods evaluated on the candidate.

\subsection{Few-Shot Adaptation to Unseen Tasks}
\label{app:fewshot_procedure}
For unseen-task generalization we treat the small support set $\mathcal{D}_i^{\text{few}}$ as a one-task version of the Stage~I data, and run a three-step adaptation procedure that exactly mirrors the standard Stage~I/II/III pipeline.

\textbf{(i) Lightweight objective adaptation.}
Starting from the Stage~I adapters, we run a short adapt phase that updates the objective-specific adapters $\{\theta_{\text{LoRA}}^{\text{O}_j}\}_{j=1}^{6}$ together with the shared adapter $\theta_{\text{LoRA}}^{\text{sh}}$ on $\mathcal{D}_i^{\text{few}}$, while keeping the LLM backbone $\theta_{\text{base}}$ frozen. Each objective head therefore better reflects the dynamics, rewards, and semantics of the new task~$i$.

\textbf{(ii) Multi-objective generation and filtering.}
We then use the adapted $\text{O}_3$ adapter to generate synthetic rollouts. For each initial state--action pair $(s_0,a_0)$ sampled from $\mathcal{D}_i^{\text{few}}$, we draw $N$ candidate trajectories $\{\tilde{\tau}^{(m)}\}_{m=1}^N$ by autoregressively rolling out the policy conditioned on the task description. Each candidate is scored under all six adapted objectives ($\text{O}_1$--$\text{O}_6$) and ranked by the aggregate score $S(\tilde{\tau}^{(m)})$; we retain the top-$K$ candidates per seed pair to form a high-quality synthetic set $\mathcal{D}_i^{\text{syn}}$.

\textbf{(iii) Policy learning on the union.}
Finally, we run Stage~III on the union $\mathcal{D}_i^{\text{few}} \cup \mathcal{D}_i^{\text{syn}}$, exactly as in Algorithm~\ref{alg:stage3}: we initialize the policy adapter $\theta_{\text{LoRA}}^{\pi} \leftarrow \theta_{\text{LoRA}}^{\text{sh}}$ (using the just-adapted shared adapter), and update $\theta_{\text{LoRA}}^{\pi}$ on this union with $\theta_{\text{base}}$ frozen. This yields an adapted policy for task~$i$ that has never interacted with the environment online; we evaluate it directly under the standard 50-episode protocol.

\section{Environment Details}
\label{app:environments}

\subsection{Meta-World}
The Meta-World benchmark~\citep{yu2020meta} is a comprehensive suite of robotic manipulation tasks designed to evaluate generalization and multi-task learning in continuous control.
It involves a Sawyer robotic arm performing 50 distinct manipulation tasks (MT50), each characterized by unique object geometries, joint configurations, and goal dynamics.
The robot operates in a 4-dimensional fine-grained continuous action space, controlling the 3D end-effector position and gripper openness.
Following recent advances in multi-task RL~\citep{yang2020multi, hu2024harmodt}, we employ the random-goal variant (MT50-rand), where goal positions are randomized per episode to promote adaptability and discourage overfitting to fixed targets.
In our experiments, we focus on a subset of 30 tasks (ML30) to balance computational efficiency with comprehensive evaluation across diverse manipulation skills.
The primary evaluation metric is the average success rate across tasks, reflecting the robot's capacity to generalize under varied and unseen goal configurations.

For offline data collection, we adopt the Soft Actor-Critic (SAC) algorithm~\citep{haarnoja2018soft} to train individual policies for each task until convergence.
For every task, we record 1M transitions during training, resulting in a comprehensive offline dataset composed of both successful and failed attempts.
Based on the convergence stage, we curate two datasets matching the main-text definitions:
\begin{itemize}
    \item \textbf{Near-Optimal (NO):} the complete 1M transitions per task collected from the converged SAC policy, dominated by expert behavior.
    \item \textbf{Sub-Optimal (SO):} a 1M-transition mixture of (i) 0.5M expert transitions sampled from the converged SAC policy and (ii) 0.5M non-expert transitions collected from a mid-performance SAC policy, yielding a 50\% expert / 50\% non-expert composition consistent with the main text.
\end{itemize}

These datasets jointly provide a rich distribution of state--action pairs spanning skill acquisition stages, supporting both imitation and reinforcement learning settings under low-data constraints.

\subsection{MuJoCo}
We further evaluate on MuJoCo continuous-control benchmarks~\citep{todorov2012mujoco} that require generalization across goal-conditioned tasks, including Cheetah-dir, Cheetah-vel, and Ant-dir.

\paragraph{Dataset construction.}
For MuJoCo tasks, we follow the standard offline data generation protocol in prior offline RL work~\citep{Mitchell2020OfflineML}, where trajectories are collected by training a policy with SAC~\citep{haarnoja2018soft} and applying action penalties to discourage excessively large control inputs.
We consider two dataset expertise levels matching the main-text definitions:
\begin{itemize}
    \item \textbf{Near-Optimal (NO):} trajectories generated by a converged expert SAC/TD3 policy, dominated by expert behavior.
    \item \textbf{Sub-Optimal (SO):} a mixture of (i) 50\% trajectories sampled from the NO dataset and (ii) 50\% trajectories collected from mid-performance SAC policies, resulting in a 50\% expert / 50\% non-expert composition with comparable data volume.
\end{itemize}
These two regimes allow us to systematically evaluate robustness to trajectory quality while keeping the overall data protocol consistent.

\paragraph{Task families.}
We evaluate across three MuJoCo benchmarks:
\begin{itemize}
    \item \textbf{Cheetah-dir:} a HalfCheetah agent is conditioned on a discrete direction variable indicating whether it should run forward or backward, with a reward encouraging movement along the specified direction. In the \emph{multi-task offline RL} setting, training and evaluation are performed on the two direction tasks. In the \emph{meta offline RL} setting, the agent is trained on one direction and evaluated on the held-out opposite direction.
    \item \textbf{Cheetah-vel:} the agent is conditioned on a target forward velocity and receives a velocity-tracking reward that penalizes deviation from the specified speed. In the \emph{multi-task offline RL} setting, the agent is trained and evaluated on all 40 velocity tasks. In the \emph{meta offline RL} setting, training is conducted on 35 target velocities and evaluation is performed on the remaining 5 unseen velocities.
    \item \textbf{Ant-dir:} the Ant agent is conditioned on a target movement direction in the 2D plane and rewarded for moving toward that direction. In the \emph{multi-task offline RL} setting, all 50 direction tasks are used for both training and evaluation. In the \emph{meta offline RL} setting, the agent is trained on 45 directions and evaluated on 5 held-out unseen directions.
\end{itemize}
Overall, these MuJoCo benchmarks complement Meta-World by emphasizing locomotion control and goal-conditioned dynamics, enabling a comprehensive evaluation of LaMA-DT across heterogeneous offline control domains under both NO and SO dataset expertise.

\section{Baselines}
\label{app:baselines}

We compare \textbf{LaMA-DT} against representative \emph{multi-task offline RL} baselines spanning value-based, behavior-cloning, generative, and Decision-Transformer paradigms.
Across all baselines, we use the same task splits and dataset preprocessing as in our main experiments, and we follow the original papers' recommended architectures and hyperparameters whenever available.

\paragraph{CDS (Conservative Data Sharing)~\citep{yu2021conservative}.}
CDS is a representative \emph{data sharing} baseline for multi-task offline RL. Its key idea is to selectively share transitions across tasks while remaining conservative under distribution shift.
In our implementation, we train a single task-conditioned actor $\pi_\theta(a\mid s,i)$ that takes the state $s$ and a task identifier $i$ as input, and outputs the action.

\paragraph{MTBC (Multi-Task Behavior Cloning).}
We extend behavior cloning to multi-task offline policy learning by using a task-ID conditioned actor that learns a single policy across all tasks, minimizing the MSE for continuous actions.

\paragraph{MTIQL (Multi-Task Implicit Q-Learning).}
IQL~\citep{IQL} learns the value function via expectile regression and trains the actor via advantage-weighted regression, which makes it effective for offline settings without behavior policy constraints as strict as CQL-style methods. MTIQL extends IQL to multi-task learning by using a task-ID conditioned actor and multi-head critic networks.

\paragraph{MTDIFF~\citep{he2023diffusion}.}
MTDIFF is a diffusion-based approach for multi-task RL that uses conditional generative modeling for planning and data synthesis. We use MTDIFF-P as our baseline.

\paragraph{MTDT (Multi-Task Decision Transformer)~\citep{chen2021decision}.}
MTDT extends Decision Transformer to multi-task data by conditioning the Transformer policy on a task identifier embedding.
Trajectories are serialized into token sequences and the model is trained with an autoregressive action prediction objective. At evaluation, MTDT is provided with the task ID and desired return-to-go to generate task-specific actions.

\paragraph{Prompt-DT (PDT)~\citep{xu2022prompting}.}
Prompt-DT conditions action generation on \emph{trajectory prompts} rather than explicit task IDs. A small set of prompt trajectories is prepended to the input sequence, serving as in-context demonstrations of task behavior. The model then predicts actions for the query trajectory conditioned on both the prompt tokens and return-to-go.

\paragraph{HarmoDT~\citep{hu2024harmodt}.}
HarmoDT is a prompt-based Decision Transformer baseline for multi-task offline RL that mitigates \emph{gradient conflicts} by learning \emph{task-specific parameter masks}, enabling conflict-aware sharing across tasks. We use \textbf{HarmoDT-F} as the HarmoDT baseline.

\paragraph{Low-data setting.}
To evaluate data efficiency, we construct low-data regimes by subsampling each task's offline dataset to a fixed percentage of the original size (e.g., 1\%, 5\%, 10\%).
Crucially, we preserve trajectory boundaries during subsampling: instead of randomly shuffling individual transitions across episodes, we sample \emph{complete trajectories} from the original dataset.
This ensures that all sequence-modeling baselines receive consistent temporal structure, and that our six objectives can be trained properly.
While the subsampling percentage is defined with respect to the \emph{number of transitions}, the actual sampling unit remains a \emph{full trajectory} (i.e., we include all transitions within each selected episode). For each task, we randomly sample trajectories without replacement until the target percentage is reached, maintaining the same trajectory-level statistics as the full dataset, only scaled down in volume.

\section{Implementation Details and Hyper-parameters}
\label{app:implementation}

We document the implementation that produced the headline results in Tab.~\ref{tab:main_results}--Tab.~\ref{tab:lowdata}. Wherever an implementation choice could deviate from the description in Sec.~\ref{sec:augmentation}--Sec.~\ref{sec:pdt}, we follow the main paper. Anything below should therefore be read as concrete realizations of the abstractions in the method, not as additional design.

\subsection{Backbone, LoRA, and Numeric Tokenization}
\textbf{Backbone.} The frozen LLM is Llama-3.2-1B with $16$ transformer blocks, hidden size $2048$, $32$ attention heads, and SiLU activations in the feedforward blocks. All base parameters have \texttt{requires\_grad=False} throughout.

\textbf{LoRA placement.} Each LoRA adapter is attached only to the attention output projection (\texttt{o\_proj}) and the MLP down-projection (\texttt{down\_proj}) of every transformer block, with rank $r=16$, scaling $\alpha=32$ (effective $\alpha/r = 2$), and dropout $0.05$ for Cheetah-dir / Cheetah-vel / Ant-dir, $0.10$ for Meta-World ML30. This placement keeps the trainable footprint small while still injecting capacity into both the attention and MLP paths.\footnote{The method only requires that $\theta_{\text{LoRA}}^{\text{O}_j}$ and $\theta_{\text{LoRA}}^{\text{sh}}$ are objective-specific / shared adapters; the choice of which two projections to attach LoRA to is an implementation detail and does not change Sec.~\ref{sec:finetune}.}

\textbf{Numeric tokenization.} Continuous states, actions, and per-step return-to-go are mapped into the LLM hidden space by three independent one-layer MLPs:

\[
\begin{aligned}
\mathtt{state\_mlp}(s)
  &= \tanh(W_s s + b_s), \\
\mathtt{action\_mlp}(a)
  &= \tanh(W_a a + b_a), \\
\mathtt{return\_mlp}(\hat{R})
  &= \tanh(W_r \hat{R} + b_r).
\end{aligned}
\]

each producing a $2048$-dim token. Continuous predictions are read out by symmetric one-layer linear heads (\texttt{state\_head}, \texttt{action\_head}, \texttt{reward\_head}) without an output non-linearity.

\textbf{Sequence layout.} A training input is $[\,\texttt{Emb}(\mathbf{p}_i,z_{\text{prov}});\,\hat{R}_0, s_0, a_0;\,\hat{R}_1, s_1, a_1;\,\ldots\,]$ with a context window of $L=20$ time steps (i.e., $60$ numeric positions plus the text prefix). Trajectories shorter than $L$ are right-padded by repeating the last state and zero-padding the action and return tokens. State and return-to-go are normalized per task using statistics computed on the real offline split.

\subsection{Stage~I Setup (Multi-Objective LoRA Fine-Tuning)}
At every step, one objective $j \sim \mathrm{Uniform}(\{1,\ldots,6\})$ and one task $i \sim \mathrm{Uniform}(\{1,\ldots,M\})$ are drawn; the corresponding $(x_j, y_j)$ is constructed from a randomly sliced window. Only $\theta_{\text{LoRA}}^{\text{O}_j}$ and $\theta_{\text{LoRA}}^{\text{sh}}$ are updated. We use AdamW (learning rate $1\times 10^{-4}$, weight decay $0.01$), gradient accumulation over four micro-batches, gradient clipping at $\|\nabla\|_2 \le 1.0$, and a linear warmup ($1{,}000$ steps for MuJoCo, $2{,}000$ for Meta-World) followed by a cosine decay. The total number of optimization steps is $50{,}000$ for the MuJoCo configurations and $100{,}000$ for Meta-World ML30. Stage~I uses bf16 mixed precision with gradient checkpointing, and is run under DDP across two NVIDIA A6000 GPUs.

\subsection{Stage~II Setup (Augmentation Generation and Filtering)}
We sample one initial $(s_0, a_0)$ per real trajectory and generate $N=4$ candidate rollouts under the $\text{O}_3$ adapter at sampling temperature $\mathcal{T}=0.4$ (chosen by the ablation in Tab.~\ref{tab:temp_sensitivity}). Long horizons are produced via chunked rollouts of length $50$ that re-anchor on intermediate teacher-forced states; this preserves the seed and final state and reduces wall-clock time by roughly $3\times$ relative to fully autoregressive generation. Each candidate is then scored under all six objectives by reactivating the relevant LoRA and running a single forward teacher-forcing pass; per-objective scores are length-normalized and combined by an unweighted mean as in Sec.~\ref{sec:pdt}. We retain the top-$1$ candidate per seed pair (i.e., $K=1$), tag it with $z_{\text{prov}}=\texttt{augmented}$, and write the resulting per-task augmented set to disk.

\subsection{Stage~III Setup (Multi-Task Policy Learning)}
The policy adapter $\theta_{\text{LoRA}}^{\pi}$ is initialized from $\theta_{\text{LoRA}}^{\text{sh}}$; the numeric heads are initialized from scratch. Each batch mixes real and augmented trajectories at the augmentation ratio $r=|\mathcal{D}_i^{\text{aug}}|/|\mathcal{D}_i|$ (default $r=0.5$, swept in Fig.~\ref{fig:ratio}). For Meta-World ML30 we additionally enable balanced sampling (a \texttt{WeightedRandomSampler} that gives every task equal frequency in expectation) to compensate for unequal per-task trajectory counts. The loss is the action MSE in Eq.~\ref{eq:lama_dt_loss}; reward and state heads are not supervised at this stage. Optimization uses AdamW with learning rate $5\times 10^{-5}$ (MuJoCo) or $1\times 10^{-4}$ (Meta-World ML30), weight decay $0.01$ (MuJoCo) or $1\times 10^{-4}$ (Meta-World), warmup over $2{,}000$--$3{,}000$ steps, and a cosine decay over the full schedule of $100{,}000$ (MuJoCo) and (Meta-World) steps. Gradient clipping is $\|\nabla\|_2 \le 1.0$ for MuJoCo and the tighter $\|\nabla\|_2 \le 0.25$ for Meta-World ML30, which we found necessary for stable late-stage optimization on the heterogeneous task suite. The training batch size is $32$.

\subsection{Evaluation Protocol}
Final policies are evaluated for $50$ episodes per task, averaged over three random seeds for Tab.~\ref{tab:main_results} and Tab.~\ref{tab:lowdata}. The provenance token is fixed to \texttt{real} at deployment. The return-to-go is initialized to the target value listed in Tab.~\ref{tab:rtg_targets} and decremented by the actual environment reward at each step. We report average normalized return for MuJoCo and per-task average success rate for Meta-World, following the conventions of prior prompt-based DT work~\citep{xu2022prompting,hu2024harmodt}.

\begin{table}[H]
\centering
\caption{Per-environment evaluation settings used to obtain Tab.~\ref{tab:main_results} and Tab.~\ref{tab:lowdata}.}
\label{tab:rtg_targets}
\setlength{\tabcolsep}{3.5pt}
\renewcommand{\arraystretch}{1.05}
\small
\begin{tabular}{l|cccc}
\toprule
& \textbf{Cheetah-dir} & \textbf{Cheetah-vel} & \textbf{Ant-dir} & \textbf{Meta-World ML30} \\
\midrule
Episodes / task                       & 50 & 50 & 50 & 50 \\
RTG target (\texttt{forward} task)    & 760 & per-task ref. velocity & per-task ref. dir. & 4{,}500 \\
RTG normalization                     & yes & yes & yes & yes \\
Provenance at test time               & \texttt{real} & \texttt{real} & \texttt{real} & \texttt{real} \\
\bottomrule
\end{tabular}
\end{table}

\subsection{Augmentation Quality Metrics: Definitions and Computation}
\label{app:aug_quality_metrics}

We define the four columns of Tab.~\ref{tab:aug_quality}. Each trajectory is flattened into one-step transitions $(s_t, a_t, s_{t+1})$, and we work in $z$-score-normalized state space, with per-dimension mean and standard deviation estimated on $\mathcal{D}^{\text{real}}$ and applied to both sets.

\paragraph{Dyn.\ MSE (one-step dynamics fidelity, $\downarrow$).}
For every augmented transition we reset the MuJoCo simulator to $s_t$, apply $a_t$, and read off the true next state $f^\star(s_t,a_t)$. Dyn.\ MSE is the per-dimension squared error between the augmented next state and $f^\star(s_t,a_t)$ in normalized space, averaged over all transitions in $\mathcal{D}^{\text{aug}}$. Real data is generated by $f^\star$ itself, so its residual is floating-point noise (rounded to $0.0000$).

\paragraph{State KL vs.\ real ($\downarrow$).}
We estimate $\mathrm{KL}(P_{\text{aug}} \,\|\, P_{\text{real}})$ between the empirical state distributions using a non-parametric $k$-NN estimator with $k=5$ and a \texttt{KDTree} per set, which avoids density assumptions in high-dimensional state space. Real-vs-real is $0$ by construction.

\paragraph{L2 Dist.\ from Dataset.}
The average $\ell_2$ distance from each augmented state to its nearest real state in normalized space (computed via a $k$-d tree). This is a proximity rather than distributional measure: both memorization (too small) and off-manifold drift (too large) are undesirable, so it is reported without a directional arrow and is meaningful only against a baseline (here MTDIFF). Real-vs-real is omitted because every real state is its own nearest neighbor.

\paragraph{Valid Coverage ($\uparrow$).}
We project all $z$-scored states onto the top-2 PCA directions fitted on $\mathcal{D}^{\text{real}}$ and discretize the resulting plane into a uniform $40\!\times\!40$ grid over the real bounding box. A transition is \emph{valid} if its per-dimension Dyn.\ MSE falls within $5\times$ the median Dyn.\ MSE on a held-out slice of $\mathcal{D}^{\text{real}}$ (admitting genuine simulator residuals while rejecting physics violations). Valid Coverage is the percentage of real-occupied grid cells matched by at least one valid augmented transition, where additional cells outside the real region but containing valid transitions push the score above $100\%$. Thus $<\!100\%$ leaves part of the real region uncovered; $100\%$ means full in-region coverage; $>\!100\%$ indicates full real coverage plus dynamics-validated novelty (e.g., the filtered $109\%$ reflects $9\%$ validated novelty beyond the real envelope).

\subsection{Hyper-parameter Summary}

Table~\ref{tab:lamadt_hparams_full} consolidates the per-stage settings actually used to obtain the headline numbers. Where two values appear (e.g.,\ ``$0.05$ / $0.10$''), they correspond to MuJoCo / Meta-World respectively.

\begin{table}[!htbp]
\centering
\caption{Hyper-parameters of LaMA-DT in our experiments. All values reflect the configuration that produced Tab.~\ref{tab:main_results}.}
\label{tab:lamadt_hparams_full}
\setlength{\tabcolsep}{6pt}
\renewcommand{\arraystretch}{1.05}
\small
\begin{tabular}{l|c}
\toprule
Parameter & Value \\
\midrule
\multicolumn{2}{l}{\textit{Backbone \& tokenization}} \\
Base LLM backbone                                        & Llama 3.2 1B (frozen) \\
Number of layers / attention heads                       & 16 / 32 \\
Hidden size                                              & 2048 \\
Activation (FFN)                                         & SiLU \\
Numeric MLPs (state / action / return)                   & 1 layer, Tanh, $\to 2048$ \\
Numeric heads (state / action / reward)                  & 1 layer, linear \\
Context window $L$                                       & 20 time steps \\
\midrule
\multicolumn{2}{l}{\textit{LoRA configuration (all stages)}} \\
LoRA target modules                                      & \texttt{o\_proj}, \texttt{down\_proj} \\
Rank $r$                                                 & 16 \\
Scaling $\alpha$                                         & 32 \\
Dropout (MuJoCo / Meta-World)                            & $0.05$ / $0.10$ \\
\midrule
\multicolumn{2}{l}{\textit{Stage~I --- multi-objective fine-tuning}} \\
Total optimization steps (MuJoCo / Meta-World)           & $50{,}000$ / $100{,}000$ \\
Optimizer                                                & AdamW, lr $1\times 10^{-4}$, wd $0.01$ \\
Warmup / decay                                           & linear $1{,}000$--$2{,}000$ / cosine \\
Gradient accumulation / clipping                         & 4 / $\|\nabla\|_2 \le 1.0$ \\
Per-step objective sampling                              & $\mathrm{Uniform}(\{1,\ldots,6\})$ \\
Mixed precision / grad checkpointing                     & bf16 / on \\
DDP                                                      & 2 GPUs (Stage~I only) \\
\midrule
\multicolumn{2}{l}{\textit{Stage~II --- augmentation \& filtering}} \\
Candidates per seed $(s_0, a_0)$, $N$                    & 4 \\
Top-$K$ retained per seed                                & 1 \\
Sampling temperature $\mathcal{T}$                       & 0.4 \\
Generation chunk length                                  & 50 steps (chunked autoregressive) \\
Score aggregation                                        & unweighted mean of $S_{\text{O}_1},\ldots,S_{\text{O}_6}$ \\
\midrule
\multicolumn{2}{l}{\textit{Stage~III --- policy learning}} \\
Total optimization steps (MuJoCo / Meta-World)           & $100{,}000$ / $200{,}000$ \\
Optimizer                                                & AdamW \\
Learning rate (MuJoCo / Meta-World)                      & $5\times 10^{-5}$ / $1\times 10^{-4}$ \\
Weight decay (MuJoCo / Meta-World)                       & $0.01$ / $1\times 10^{-4}$ \\
Warmup / decay                                           & $2{,}000$--$3{,}000$ / cosine \\
Gradient clipping (MuJoCo / Meta-World)                  & $\|\nabla\|_2 \le 1.0$ / $\|\nabla\|_2 \le 0.25$ \\
Batch size                                               & 32 \\
Augmentation ratio $r$                                   & $0.5$ (default; swept in Fig.~\ref{fig:ratio}) \\

Loss                                                     & action MSE (Eq.~\ref{eq:lama_dt_loss}) \\
Policy adapter init                                      & $\theta_{\text{LoRA}}^{\pi} \leftarrow \theta_{\text{LoRA}}^{\text{sh}}$ \\
\midrule
\multicolumn{2}{l}{\textit{Evaluation}} \\
Episodes per task                                        & 50 \\
Random seeds                                             & 3 \\
Provenance token at test                                 & \texttt{real} \\
RTG initialization                                       & per-task target (Tab.~\ref{tab:rtg_targets}) \\
\bottomrule
\end{tabular}
\end{table}

\subsection{Example Task Descriptions}
\label{app:task_descriptions}
Below we provide example natural-language task descriptions $\mathbf{p}_i$ used in our experiments, following the unified template introduced in Sec.~\ref{sec:augmentation}.
\begin{itemize}
    \item \textbf{Cheetah-dir:} ``This is the Cheetah-dir task, and its goal is to run forward along the $x$-axis at high speed. The environment is the MuJoCo HalfCheetah-v2, a planar 9-DoF robot. Episodes run up to 1{,}000 steps. The trajectories show expert-level performance.''
    \item \textbf{Cheetah-vel:} ``This is the Cheetah-vel task, and its goal is to maintain a target running velocity of 2 m/s. The environment is the MuJoCo HalfCheetah-v2, a planar 9-DoF robot. Episodes run up to 1{,}000 steps. The trajectories show expert-level performance.''
    \item \textbf{Ant-dir:} ``This is the Ant-dir task, and its goal is to move toward roughly 90 degrees. The environment is the MuJoCo Ant-v2, an 8-joint quadruped ant robot. Episodes run up to 1{,}000 steps. The trajectories show expert-level performance.''
    \item \textbf{Meta-World MT50 (drawer-open):} ``This is the drawer-open task, and its goal is to use a 7-DoF Sawyer robotic arm to grasp the handle of a closed drawer and pull it open to a target displacement of more than 0.2 meters. The environment is a simulated tabletop workspace where a drawer is placed at a randomized initial position, and episodes run up to 150 steps. The trajectories show expert-level performance.''
\end{itemize}

\section{Additional Experimental Results}
\label{app:additional_results}

\subsection{Training Dynamics}
\begin{figure}[H]
\centering
\includegraphics[width=1\textwidth]{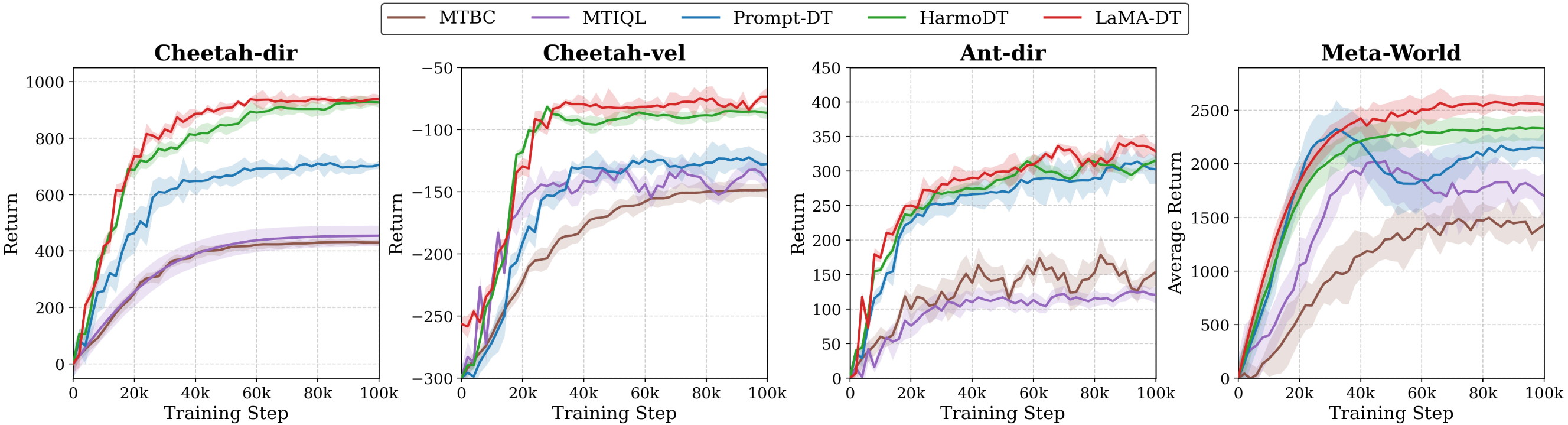}
\caption{Training curves of LaMA-DT and the baselines in Tab.~\ref{tab:main_results}.}
\label{fig:curve}
\end{figure}

Figure~\ref{fig:curve} shows the training curves of LaMA-DT and strong baselines across Cheetah-dir, Cheetah-vel, and Ant-dir.
LaMA-DT consistently achieves the highest final performance in all three environments. More notably, LaMA-DT exhibits faster convergence in the early training phase compared to Prompt-DT and HarmoDT. This accelerated early-stage learning can be attributed to the pretrained shared LoRA adapter $\theta_{\text{LoRA}}^{\text{sh}}$ from Stage~I, which has already learned task-agnostic representations of dynamics, rewards, and semantic patterns across the multi-objective objectives ($\text{O}_1$--$\text{O}_6$).
By initializing the policy adapter $\theta_{\text{LoRA}}^{\pi}$ from these shared LoRA weights rather than from random initialization, LaMA-DT starts from a more favorable point in parameter space, enabling faster adaptation to the specific decision-making task.

\subsection{Sensitivity to Generation Temperature}
\label{app:temperature}
To study how stochasticity affects trajectory augmentation quality, we vary the sampling temperature $\mathcal{T}$ when generating candidate trajectories while keeping the augmentation budget ($N,K$), scoring/filtering, and downstream policy training protocol fixed.
Table~\ref{tab:temp_sensitivity} reports performance on Meta-World ML30 (success rate) and MuJoCo Cheetah-dir (normalized return).

\begin{table}[H]
\centering
\caption{Sensitivity to generation temperature $\mathcal{T}$ in trajectory augmentation. We report Meta-World ML30 success rate (\%) and MuJoCo Cheetah-dir normalized return. Results are averaged over 3 seeds.}
\label{tab:temp_sensitivity}
\setlength{\tabcolsep}{6pt}
\renewcommand{\arraystretch}{1.08}
\small
\begin{tabular}{c|cc}
\toprule
\textbf{Temperature $\mathcal{T}$} & \textbf{ML30} & \textbf{Cheetah-dir} \\
\midrule
0.0 & 63.0$\pm$1.1 & 896.8$\pm$10.8 \\
0.2 & 67.3$\pm$1.3 & 917.3$\pm$16.7 \\
\textbf{0.4 (default)} & \textbf{69.8$\pm$1.7} & \textbf{937.8$\pm$9.6} \\
0.8 & 59.6$\pm$2.2 & 874.9$\pm$15.0 \\
\bottomrule
\end{tabular}
\end{table}

Overall, we observe a clear trade-off between \emph{diversity} and \emph{fidelity}.
Deterministic decoding ($\mathcal{T}{=}0$) yields conservative rollouts that are easier to score but lack coverage, resulting in the lowest scores on both benchmarks.
A mild temperature ($\mathcal{T}{=}0.2$--$0.4$) improves augmentation diversity while preserving task/dynamics consistency, with $\mathcal{T}{=}0.4$ achieving the best result on both ML30 and Cheetah-dir; these numbers also match the headline values in Tab.~\ref{tab:main_results}.
Overly stochastic sampling ($\mathcal{T}{=}0.8$) degrades both benchmarks, suggesting that high-temperature generations introduce off-manifold transitions that cannot be fully corrected by reranking.
We therefore use $\mathcal{T}{=}0.4$ as the default in all other experiments.

\subsection{Effect of Dataset Expertise}
\label{app:expertise}
We further evaluate all methods under two dataset expertise levels in the \emph{standard} data regime, \emph{Near-Optimal} (NO) and \emph{Sub-Optimal} (SO), on ML30 (success rate) and Cheetah-dir (normalized return).
As summarized in Table~\ref{tab:expertise}, performance decreases from NO to SO for all baselines, confirming that learning solely from lower-quality trajectories is intrinsically harder even when the overall data volume is kept fixed.
However, LaMA-DT consistently outperforms prior methods in both expertise regimes and exhibits a smaller NO$\rightarrow$SO performance drop: on ML30, LaMA-DT improves over HarmoDT by $+5.9$ and $+6.4$ points in NO and SO, respectively, while on Cheetah-dir it achieves both the highest absolute return and the most stable degradation.
Compared to the low-data regime results, where our gains are primarily attributed to better data efficiency under scarce demonstrations, these findings show that LaMA-DT also brings substantial benefits in the standard-data setting by more effectively exploiting heterogeneous-quality trajectories and mitigating the impact of sub-optimal supervision.

\begin{table}[H]
\centering
\setlength{\abovecaptionskip}{2pt}
\setlength{\belowcaptionskip}{2pt}
\caption{Performance across different dataset expertise under \textbf{standard data regime}: Near-Optimal (NO) vs.\ Sub-Optimal (SO). We report success rate for ML30 and normalized return for Cheetah-dir.}
\label{tab:expertise}
\setlength{\tabcolsep}{4pt}
\small
\begin{tabular}{l|cc|cc}
\toprule
\multirow{2}{*}{\textbf{Method}} & \multicolumn{2}{c|}{\textbf{ML30}} & \multicolumn{2}{c}{\textbf{Cheetah-dir}} \\
\cmidrule(lr){2-3}\cmidrule(lr){4-5}
 & \textbf{NO} & \textbf{SO} & \textbf{NO} & \textbf{SO} \\
\midrule
MTDT & 70.9$\pm$1.6 & 55.2$\pm$2.1 & 678.5$\pm$24.3 & 598.1$\pm$22.9 \\
PDT & 72.1$\pm$0.9 & 58.7$\pm$1.3 & 705.6$\pm$8.3 & 636.7$\pm$11.2 \\
HarmoDT & 82.7$\pm$1.2 & 63.4$\pm$1.6 & 926.3$\pm$10.4 & 872.4$\pm$13.8 \\
\midrule
\textbf{LaMA-DT} & \textbf{88.6$\pm$1.5} & \textbf{69.8$\pm$1.7} & \textbf{937.8$\pm$9.6} & \textbf{912.3$\pm$14.9} \\
\bottomrule
\end{tabular}
\end{table}

\subsection{Effect of LLM Backbone}
\label{app:backbone}

To disentangle whether LaMA-DT's gains stem from \emph{model capacity} or from \emph{pre-training quality}, we instantiate the same three-stage pipeline (Stage~I objectives, Stage~II generation/filter, Stage~III DT learning) with four backbones of different sizes and pre-training corpora: GPT-2 (124M), Qwen2.5-0.5B (494M), Qwen2.5-1.5B (1.5B), and Llama~3.2 1B (1.24B). All other hyper-parameters (LoRA rank, learning rate, candidates, augmentation ratio, normalization) are kept identical, and we evaluate on Cheetah-dir.

\begin{table}[H]
\centering
\caption{LaMA-DT performance with different LLM backbones on Cheetah-dir (50 episodes/task), averaged over the two task directions.}
\label{tab:backbone}
\setlength{\tabcolsep}{8pt}
\renewcommand{\arraystretch}{1.05}
\small
\begin{tabular}{lc|c}
\toprule
\textbf{Backbone} & \textbf{Params} & \textbf{Average Return} \\
\midrule
GPT-2                          & 124M  & 386.2 \\
Qwen2.5-0.5B                   & 494M  & 888.0 \\
Qwen2.5-1.5B                   & 1.5B  & 917.0 \\
\textbf{Llama~3.2 1B (default)} & 1.24B & \textbf{937.8} \\
\bottomrule
\end{tabular}
\end{table}

Three observations follow.
(i) \emph{Pre-training quality dominates raw size.} GPT-2 (124M) collapses on the forward task, producing a low average ($386.2$) because its older pre-training corpus does not provide the language priors required for our $\text{O}_5/\text{O}_6$ alignment objectives, whereas Qwen2.5-0.5B (less than half the size of Llama) already reaches $888.0$ ($94.7\%$ of Llama's score).
(ii) \emph{Capacity helps but with diminishing returns.} Moving from Qwen2.5-0.5B$\rightarrow$1.5B$\rightarrow$Llama~3.2~1B yields $+29.0$ and $+20.8$ on average, much smaller than the $+501.8$ gap from GPT-2$\rightarrow$Qwen2.5-0.5B at similar size, indicating gains are not explained by parameter count alone.
(iii) \emph{Method gains are separable from backbone gains.} Within the Llama backbone, our augmentation pipeline still adds $+12$ over a no-augmentation control, and the multi-objective filter adds another $+54.7$ (Tab.~\ref{tab:aug_quality}); these are method-specific rather than capacity-specific.
Together, the backbone sweep provides a controlled answer to ``is LaMA-DT just a bigger model?'': a well-pretrained sub-billion backbone already realizes most of the gains, and the rest comes from the proposed objectives, generation, and filtering rather than scale.

\subsection{Robustness to Task-Description Variations}
\label{app:description_robustness}

A natural concern is whether LaMA-DT's gains depend on specific template phrasing or, more fundamentally, on the discriminative semantic content the descriptions carry. To disentangle these, we evaluate six variants on ML30 and Cheetah-dir, keeping the rest of the pipeline identical. Table~\ref{tab:template} reveals a clear pattern: performance scales monotonically with the amount of informative content preserved.

\begin{table}[H]
\centering
\caption{Robustness to task-description variations. ML30: average success rate (\%); Cheetah-dir: average normalized return.}
\label{tab:template}
\setlength{\tabcolsep}{6pt}
\renewcommand{\arraystretch}{1.0}
\small
\begin{tabular}{l|cc}
\toprule
\textbf{Description variant} & \textbf{ML30 (succ.\ \%)} & \textbf{Cheetah-dir (return)} \\
\midrule
Full template (default)                & \textbf{69.8}          & \textbf{937.8} \\
\,\,Minimal (task name only)           & 55.5          & 892.2 \\
\,\,Drop-expertise field               & 63.0          & 920.3 \\
\,\,Noisy (typos + grammar errors)     & 49.6          & 801.7 \\
\,\,Shortened paraphrase               & 67.9          & 912.0 \\
\midrule
Learnable task-ID embedding            & 61.0          & 913.4 \\
No description (shared text)           & 40.4          & 746.9 \\
\bottomrule
\end{tabular}
\end{table}

\textbf{Information-preserving variants stay close to the default.} \emph{Shortened paraphrase}, which compresses the description while retaining all task semantics, drops only $1.9$ pts on ML30 ($69.8 \rightarrow 67.9$) and $25.8$ on Cheetah-dir, indicating that LaMA-DT does not depend on a particular wording or length.

\textbf{Information-degrading variants cause progressively larger drops.} Removing the expertise field ($-6.8$ pts on ML30), reducing the description to a task name (\emph{Minimal}, $-14.3$ pts), and injecting typos/grammatical errors (\emph{Noisy}, $-20.2$ pts) all hurt performance, with \emph{Noisy} approaching the floor set by \emph{No description}. This shows that the LLM backbone is sensitive not to surface formatting per se, but to whether the description remains \emph{parsable} as coherent task semantics---noise and over-compression both destroy that signal.

\textbf{Learnable task-ID embeddings underperform language.} Replacing natural language with a randomly initialized, end-to-end trained per-task vector reaches $61.0$ on ML30 and $913.4$ on Cheetah-dir, clearly below the full template ($69.8$ / $937.8$). Combined with the unseen-task results in Table~\ref{tab:unseen_transposed}, where LaMA-DT outperforms Prompt-DT by $+8.8$ pts on ML30, this confirms that natural language carries information that an opaque ID embedding cannot recover---both for distinguishing seen tasks during training and for generalizing to new ones at test time.

\textbf{Removing discriminative content collapses performance.} The \emph{No description} variant replaces every per-task description with a single shared placeholder, eliminating all inter-task discriminative information. This drops ML30 by $29.4$ pts ($69.8 \rightarrow 40.4$) and Cheetah-dir by $190.9$ ($937.8 \rightarrow 746.9$), confirming that LaMA-DT's gains are driven by what the descriptions \emph{communicate} rather than by the mere presence of language tokens. Without per-task semantic disambiguation, augmentation degenerates into goal-agnostic generation.

In summary, LaMA-DT is robust to surface compression and rewording, but degrades smoothly as descriptions lose discriminative content. The implication for practitioners is straightforward: any natural-language summary that clearly distinguishes one task from another suffices, but minimal task names or ungrammatical text carry too little signal for the LLM to ground generation effectively.

\subsection{Generalization to Unseen Tasks}
\label{app:unseen}
We evaluate few-shot adaptation to unseen tasks using the procedure described in Sec.~\ref{app:fewshot_procedure} and compare LaMA-DT with MTDT and Prompt-DT on four benchmarks: ML30, Cheetah-vel, Cheetah-dir, and Ant-dir.
The results in Table~\ref{tab:unseen_transposed} show that LaMA-DT consistently achieves superior performance across all unseen-task settings.
In particular, LaMA-DT attains an average reward of $216.0$ across all benchmarks, substantially outperforming Prompt-DT ($173.5$), corresponding to a relative improvement of approximately $25\%$.
These results indicate that combining natural-language task descriptions with a small amount of task-specific data enables reliable data augmentation and supports fast policy adaptation to unseen tasks.

\begin{table}[H]
\centering
\setlength{\abovecaptionskip}{2pt}
\setlength{\belowcaptionskip}{2pt}
\caption{Unseen-task generalization. Meta-World success rate (\%); MuJoCo normalized return (\%). Comparison against baselines capable of prompt-conditioning.}
\label{tab:unseen_transposed}
\setlength{\tabcolsep}{6pt}
\renewcommand{\arraystretch}{1.05}
\small
\begin{tabular}{l|c|ccc|c}
\toprule
& \textbf{Meta-World} & \multicolumn{3}{c|}{\textbf{MuJoCo}} & \\
\cmidrule(lr){2-2}\cmidrule(lr){3-5}
\textbf{Method} & \textbf{ML30 (un)} & \textbf{C-vel} & \textbf{C-dir} & \textbf{Ant-dir} & \textbf{Average} \\
\midrule
MTDT     & 35.9$\pm$4.1 & -184.2$\pm$8.1  & 429.4$\pm$19.5 & 155.2$\pm$9.5  & 109.1 \\
PDT      & 43.5$\pm$2.7 & -140.2$\pm$10.9 & 577.2$\pm$8.5  & 213.6$\pm$20.4 & 173.5 \\
\midrule
\textbf{LaMA-DT} & \textbf{52.3$\pm$3.2} & \textbf{-94.1$\pm$7.6} & \textbf{668.0$\pm$16.1} & \textbf{237.5$\pm$7.6} & \textbf{216.0} \\
\bottomrule
\end{tabular}
\end{table}

\subsection{Low-Data Component Analysis and Computational Cost}
\label{app:lowdata_cost}

\paragraph{Low-data component analysis.}
Table~\ref{tab:lowdata_components} isolates two components at 5\% of the
sub-optimal training data. Removing augmentation while retaining the
shared-LoRA policy initialization reduces performance by 161.9 on
Cheetah-dir and 9.3 percentage points on ML30. Retaining the augmented
data but randomly initializing the policy adapter results in smaller
decreases of 6.6 and 2.1, respectively. Thus, the augmentation
pipeline accounts for most of the measured gain in this low-data
setting, while shared-adapter initialization provides an additional
benefit. This comparison evaluates augmentation as a whole; it does
not separately identify the effects of increased data volume and
task-conditioned generation.

\begin{table}[!htbp]
  \centering
  \small
  \caption{Component analysis with 5\% sub-optimal training data.
  Cheetah-dir reports normalized return; ML30 reports success rate
  (\%). In the first ablation, shared-LoRA initialization is retained.
  In the second, augmented data are retained.}
  \label{tab:lowdata_components}
  \begin{tabular}{lcc}
    \hline
    Method & Cheetah-dir & ML30 \\
    \hline
    Full LaMA-DT                     & 521.5 & 39.1 \\
    w/o augmentation                 & 359.6 & 29.8 \\
    w/o shared-LoRA initialization   & 514.9 & 37.0 \\
    \hline
  \end{tabular}
\end{table}

\paragraph{Adapter count and computational cost.}
LaMA-DT uses six objective-specific LoRA adapters and one shared
adapter in Stage I, followed by one policy adapter in Stage III.
These eight adapters are shared across tasks in the main experiments;
we do not fine-tune a separate adapter for each task. The 1.24B-parameter
language-model backbone remains frozen. Approximately 25.8M parameters
are trainable in Stage I and 3.79M in policy learning.

On an NVIDIA A6000 with the Llama-3.2-1B backbone, generating four
candidate trajectories for one seed pair takes 2.66\,s. Scoring those
four candidates under all six objectives takes 1.51\,s, or approximately
36\% of the combined generation-and-scoring time. Generation and
scoring are performed offline; at deployment, the policy requires
neither step and takes approximately 17\,ms per action under the same
hardware setup.

\end{document}